\documentclass[11pt]{article}

\usepackage{acl}

\usepackage{times}
\usepackage{latexsym}
\usepackage[T1]{fontenc}
\usepackage[utf8]{inputenc}
\usepackage{microtype}
\usepackage{graphicx}
\usepackage{booktabs}
\usepackage{array}
\usepackage{amsmath}
\usepackage{amssymb}
\usepackage{xcolor}

\newcommand{\masktok}{\texttt{[MASK]}}
\newcommand{\codename}[1]{\texttt{#1}}

\title{When More Becomes Less:\\Position-Dependent Repetition Effects in Language Models}
\author{Han-yu Wang \\
  The University of Hong Kong \\
  \texttt{henry.why@connect.hku.hk}}

\begin{document}
\maketitle

\begin{abstract}
Cloze-style probes that vary how often a target token appears
implicitly assume that more copies of a target affect prediction the
same way regardless of where the readout slot sits. We show this
assumption fails. Our two-probe design holds a repeated-target
prefix fixed and varies only the readout position: the
\emph{adjacent} probe places the slot immediately after the repeated
block; the \emph{displaced} probe places it inside a fresh sentence
frame. Adjacent repetition behaves as priming intuition predicts:
$P(\text{target})$ climbs with $N$ and plateaus. Displaced repetition
produces an inverted-U: $P(\text{target})$ rises to an early peak and
then declines as more copies are added. The displaced inverted-U
shows a per-word drop with bootstrap CI excluding zero in all $13$
open-access encoder and decoder models we test, and replicates across
Spanish, Chinese, German, and French in $42$ of $42$ multilingual
cells. A six-condition causal ablation isolates the effect to exact
lexical repetition rather than length, generic redundancy, or
semantic-neighbour exposure. A frame-pragmatics control rules out an
artefact of the readout frame. Internally, per-target-token attention
falls with $N$ while the total budget assigned to the repeated block
grows in causal LMs but not in the masked LM we probe. Probes that
vary repetition count cannot treat the readout position as orthogonal
to what they measure.
\end{abstract}

\section{Introduction}
\label{sec:intro}

A common pattern in cloze-style probing studies of language models
\citep{petroni2019language, ettinger2020what} is to read the model's
prediction at a single slot, typically the next-token position
after the controlled material in the prompt. When such a probe also
varies how often a target token appears, it implicitly assumes that
the readout position is orthogonal to what repetition does: that
whatever effect more copies have on prediction is the same
regardless of where the prediction slot sits relative to the
repeated block. We show this assumption fails. The same
repeated-target prefix produces opposite curves at two readout
positions one clause apart.

\begin{figure}[!htbp]
  \centering
  \includegraphics[width=\columnwidth]{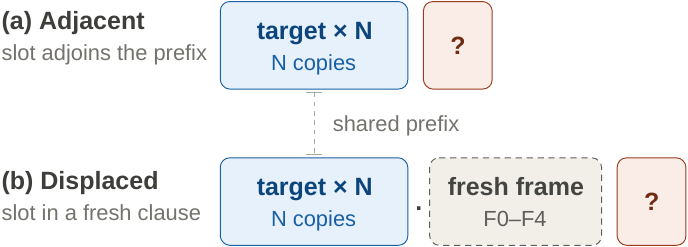}
  \caption{\textbf{Two-probe design.} (a) The \emph{adjacent} probe
    places the prediction slot directly after a prefix of $N$ target
    copies. (b) The \emph{displaced} probe shares the same prefix but
    moves the slot into a fresh clause via one of five sentence
    frames F0--F4. The framework applies identically to masked LMs
    (slot is \masktok) and to causal LMs (slot is the next-token
    position).}
  \label{fig:schematic}
\end{figure}

The diagnostic is a minimal pair (Figure~\ref{fig:schematic}). We
hold a repeated-target prefix fixed and vary only where the
prediction slot lies. The \emph{adjacent} probe places the slot
directly after the block; the \emph{displaced} probe moves it into a
fresh sentence frame. On the adjacent slot, $P(\text{target})$
climbs with the copy count $N$ and plateaus, as a priming account
predicts. On the displaced slot, $P(\text{target})$ instead rises to
an early peak and then falls as more copies are added: the shape we
call the \emph{displaced inverted-U}. Because the adjacent baseline
confirms that local continuation remains available under the same
prefix, the displaced decline cannot be read as long-context
capacity loss; it is a property of the readout position.

This dissociation matters because the displaced slot is the one that
controlled probing studies routinely score. Cloze probes of factual
knowledge, semantic-priming studies, in-context-learning analyses
that vary demonstration count, and any benchmark that reads a
prediction one clause beyond the material it manipulates all treat
the readout position as orthogonal to how the manipulation affects
prediction. The displaced inverted-U says this is wrong whenever the
manipulation is repetition: the monotone priming a study expects can
reverse at the slot it actually scores.

The effect is not a fragile corner case. It appears in every one of
the $13$ open-access encoder and decoder models we probe and
replicates across Spanish, Chinese, German, and French. A
six-condition causal ablation pins it to exact lexical repetition of
the target rather than sequence length, generic redundancy, or
exposure to the target's semantic neighbours, and a frame-pragmatics
control shows the readout frame itself is not responsible. The signature
even survives when the answer is stated explicitly at the start of
the prompt, an alternative retrieval route the displaced probe
otherwise withholds. Internally, the per-target-token attention each
repeated copy receives falls with $N$, while the total attention
budget the model assigns to the repeated block grows in the causal
LMs we examine but not in the masked LM.

Our primary contribution is the empirical finding that current
open-access models fail to aggregate exact lexical repetition
additively at displaced slots: adding copies of a target can lower
its probability one clause beyond the block, even as the same prefix
primes the adjacent slot. The second contribution is the instrument
that makes this finding visible, a matched-prefix two-probe design
that separates local pattern continuation from displaced retrieval
and applies to any controlled probing study of repeated content. The
rest of the paper establishes this finding and its boundaries.
Section~\ref{sec:reversal} documents the adjacent/displaced
dissociation across the panel and its cross-lingual replication;
Section~\ref{sec:cause} isolates the cause through a six-condition
ablation, a frame-pragmatics control, and a pre-stated-answer
robustness check; Section~\ref{sec:mech} reports the internal
attention correlate; and Section~\ref{sec:discussion} draws out the
consequences for probing methodology.

\section{Related Work}
\label{sec:related}

Most prior work on repetition in language models has been about the
model's output: degenerate loops under greedy or beam decoding
\citep{holtzman2020curious, welleck2020unlikelihood, xu2022learning},
and the role of multiple demonstrations in in-context learning
\citep{min2022rethinking, agarwal2024manyshot}. We are asking about
the input side instead: how the model's predictions change when
the prompt is repetitive.

Several lines bear on what repeated tokens do to Transformer-based
language models on the input side. \citet{barbero2024transformers}
prove a representational-collapse result for repeated decoder-only
inputs; \citet{yona2025interpreting} characterise a
``repeated-token divergence'' phenomenon and link it to
attention-sink dynamics. Attention sinks themselves
\citep{xiao2024streaming, gu2024when} have been characterised as
positions that accumulate disproportionate attention mass. Our
block-budget growth shares the flavour of the attention-sink line,
but appears under a different setup (a span at the source side
under displaced readout, rather than initial tokens carried across
all queries), and we stop short of claiming it is the same
mechanism. None of this prior work asks our question, which is
whether the \emph{position of the readout slot relative to the
repeated block} changes what repetition does to retrieval. The
experimental object here is a behavioural dissociation between two
probe positions under a matched prefix.

Other prompt-side phenomena include ``lost in the middle'' effects
under long prompts \citep{liu2024lostmiddle} and prompt-format and
prompt-order sensitivity in in-context-learning benchmarks
\citep{lu2022fantastically, sclar2024quantifying}.
Retrieval-augmented generation and open-domain QA pipelines are a
further source of input-side repetition: they concatenate passages
that share an entity, an answer string, or a source document,
inviting exact and near-exact copies
\citep{lewis2020rag, karpukhin2020dpr, izacard2021fid}.

Methodologically our paper is closest to the cloze-probe tradition
\citep{petroni2019language, ettinger2020what, misra2020exploring,
hu2023prompting}, but with a \emph{displaced} slot that dissociates
surface continuation from across-clause retrieval, and with the
framework extended to causal LMs by replacing \masktok\ with the
next-token position. The repetition schedule in our prefix is
conceptually inspired by the psycholinguistic literature on semantic
satiation \citep{smith1984semantic, black2001semantic,
tian2010testing}, but we do not claim a human-fatigue mechanism.

\section{Experimental Design}
\label{sec:setup}

\subsection{Two-probe framework}

Both probes use the same $N$ grid $N \in \{0, 1, 3, 5, 10, 20, 30\}$
and a shared repeated-target prefix; only the readout slot differs.
($N{=}0$ is the prior-only baseline; all reported statistics use
$N \geq 1$.)
The \emph{adjacent} probe is \codename{<target $\times$ N> [MASK] .}
for masked-LM models (MLM) or \codename{<target $\times$ N>} for
causal-LM models (CLM) at the next-token slot. The \emph{displaced}
probe is \codename{<target $\times$ N> . <frame> [MASK] .} (MLM) or
\codename{<target $\times$ N> . <frame>} (CLM). We test five English
declarative-style frames F0--F4 (Appendix~\ref{app:wordlist}).
The adjacent probe is evaluated on a three-model baseline
(ModernBERT-base, Qwen2.5-1.5B, Qwen2.5-7B) as a sanity check that
local continuation remains available. The displaced grid is
evaluated on every panel model.

A short vocabulary note collects the terms we reuse. \emph{Displaced
retrieval} names the readout condition just defined; \emph{adjacent
retrieval} names the matched local-continuation control. The
\emph{displaced inverted-U} is the rise-then-fall shape of
$P(\text{target})$ against $N$ at the displaced slot;
\emph{non-additive aggregation} is the property it implies, that exact
copies do not add monotonically to target probability one clause
beyond the block. The label \textit{satiation-like} is reserved for
per-word curves that clear the deterministic thresholds of
Appendix~\ref{app:stats}. All three describe curve shape at the
displaced slot, not internal mechanism.

\subsection{Targets and tokenisation}

The English target set is $261$ hand-curated single-token noun
candidates balanced over concreteness and Zipf frequency
\citep{speer2022wordfreq}; per-tokeniser filtering yields per-model
sets of $182$--$258$ words and a shared core of $\approx 168$ words
that enters the cross-model regression of \S\ref{sec:stat}. Construction details (cell
balance, semantic-class breakdown, per-family overlap) are in
Appendix~\ref{app:wordlist}.

\subsection{Open-access panel}
\label{sec:panel}

We probe $13$ open-access language models, all loadable without
authentication. Two encoders use the masked-LM objective
(ModernBERT-base, ModernBERT-large \citep{warner2024modernbert}) and
eleven decoders use the causal-LM objective. The Qwen2.5 family
\citep{qwen2025qwen25} provides a within-family scale ladder ($0.5$B,
$1.5$B, $3$B, $7$B, $14$B, plus the $7$B-Instruct variant).
Cross-family checks include SmolLM2 at $360$M and $1.7$B
\citep{allal2025smollm2}, Phi-3.5-mini at $3.8$B
\citep{abdin2024phi3}, and OLMo-2 at $1$B and $7$B
\citep{olmo2025olmo2}. Hardware, dtype, and prompt-format details
(including evaluating instruction-tuned variants on the raw prompt
rather than the chat template) are in Appendix~\ref{app:repro}.

\subsection{Metrics and statistics}

Our \emph{headline statistic} is the per-word drop
$(\text{peak}_{w} - P(N{=}30)_{w})/\text{peak}_{w}$ with the peak
taken over $N \in \{1,3,5,10,20,30\}$, bootstrapped at the word
level; throughout, ``the per-word drop is positive'' means the
bootstrap CI excludes zero. Alongside it we report mean-curve
inversion before $N{=}30$ as a \emph{secondary descriptive
observation}; it is more sensitive to peak location and so is
weaker evidence than the per-word drop. We also run an OLS quadratic
of $\log P(\text{target})$ on $N$, $N^{2}$, Zipf frequency, model,
frame, and word category, with SEs clustered by target word; signs
and significance of $\beta_{N}$ and $\beta_{N^{2}}$ are the
cross-model summary. For ablations and the
mechanism probes we also record the target's rank and
$\Sigma P(\text{synonyms})$ over a curated semantic field. A separate deterministic
classifier labels per-(model, frame, word) curves as
\textit{satiation-like}, \textit{monotonic-priming},
\textit{plateau}, or \textit{mixed}; thresholds and the full
regression are in Appendix~\ref{app:stats}.

\section{Probe Position Reverses Repetition Dynamics}
\label{sec:reversal}

\subsection{Adjacent vs displaced: the dissociation}
\label{sec:adjacent}

\begin{figure*}[!htbp]
  \centering
  \includegraphics[width=\textwidth]{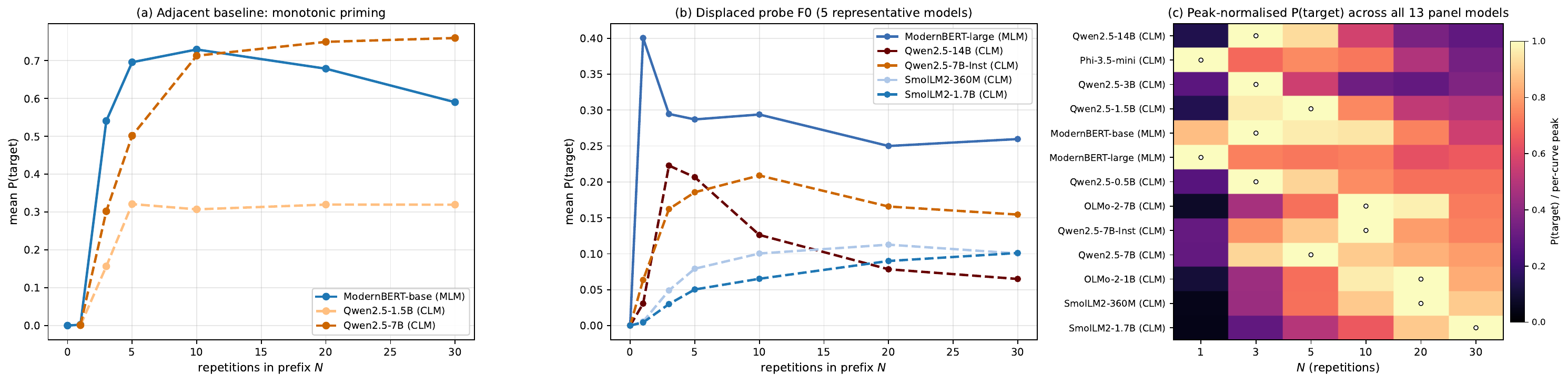}
  \caption{\textbf{Same prefix, different probe position.}
    (a) Adjacent baseline on three models: priming, plateau by
    $N{\approx}20$.
    (b) Displaced probe (F0), five representative models spanning
    the panel's peak-$N$ range. Solid: MLM; dashed: CLM.
    (c) Peak-normalised $P(\text{target})$ across all 13 panel
    models, rows sorted by F0 drop magnitude; white dot marks each
    row's peak $N$. Per-model numbers in
    Table~\ref{tab:model_results}.}
  \label{fig:main_curves}
\end{figure*}

On the adjacent slot the three baseline models behave broadly as a
priming account would predict.
$P(\text{target})$ climbs with $N$ and plateaus by $N \approx 20$
(Figure~\ref{fig:main_curves}a), and the MLM never registers a
single \textit{satiation-like} word. Move the slot off the
repeated block (same prefix, different probe position) and the
curve no longer climbs. It rises through an early peak and falls
back by $N{=}30$ (Figure~\ref{fig:main_curves}b,c), with per-word
drops clearing zero in all $13$ panel models (\S\ref{sec:dense}).
Because the adjacent baseline shows
that local continuation remains available, the displaced
decline cannot be read as generic capacity loss; it has to be
about the slot. This is the displaced inverted-U, the non-additive
aggregation defined in \S\ref{sec:setup}: adding exact copies of the
target does not monotonically add to its probability at a readout one
clause beyond the block.

\subsection{Panel-wide displaced result}
\label{sec:dense}

F0 drops on the full panel range from $0.138$ $[0.114, 0.164]$ in
SmolLM2-1.7B to $0.733$ $[0.706, 0.761]$ in Qwen2.5-14B
(Table~\ref{tab:model_results}). Every per-word drop has a bootstrap
interval that clears zero. This is our primary result: the
displaced inverted-U is a $13$-of-$13$ effect under the per-word
statistic. The five-fold spread across the
panel is large in magnitude but uniform in sign: every model
shows the inverted-U, only the strength varies, and the
ranking-by-magnitude (Phi-3.5-mini and Qwen2.5-14B at the top,
SmolLM2 at the bottom) does not align cleanly with parameter count.

As a secondary descriptive observation, $12$ of $13$ mean curves
also turn over before $N{=}30$, with per-model peaks at
$N \in \{1, 3, 5, 10, 20\}$; the exception is SmolLM2-1.7B, whose
per-word drops remain positive ($0.138$ $[0.114, 0.164]$) but whose
population-mean curve continues to climb. We treat the per-word
statistic as primary because it does not assume a single peak
location across heterogeneous targets: a curve that peaks at $N{=}3$
and a curve that peaks at $N{=}20$ contribute equally to the
per-word drop, but only the early-peak curve registers as an
inversion under any single-peak summary.

\paragraph{Frame heterogeneity.}
Across F0--F4 the per-word drop is reliably positive in all $65$ of
$65$ (model, frame) cells. The $13$-model mean drop is also
tightly clustered (F0 $0.460$, F1 $0.451$, F2 $0.483$, F3 $0.458$,
F4 $0.434$), so the main result does not hinge on which frame we
single out. F0 is among the larger frames, but no single frame
dominates. The $\approx 5$ percentage-point
spread across frames is small next to the $\approx 60$
percentage-point spread across models on F0, so frame phrasing is a
much weaker source of variance than model architecture.

\paragraph{Panel-level regression check.}
\label{sec:stat}
A cross-model OLS quadratic on shared-core cells reproduces the
inverted-U at the panel level: $\beta_{N} = +0.212$ and
$\beta_{N^{2}} = -0.00610$ ($|t|>36$ under SEs clustered by target
word, implied vertex $N^{*} \approx 17.4$).\footnote{An
average-trajectory vertex, not any individual model's peak.} A
Zipf-confound regression finds a small reliable frequency effect on
per-word drop, but (model, frame) fixed effects out-explain it by
$\approx 60\times$: frequency modulates, it does not drive. Full
coefficient tables, the cluster-robust panel regression, and the
deterministic curve classifier underlying the
\textit{satiation-like} share are in Appendix~\ref{app:stats},
Tables~\ref{tab:ols_full}--\ref{tab:freq_reg}.

\paragraph{Scale heterogeneity.}
\label{sec:scale}
Scale within the Qwen2.5 family is not a monotone driver of
magnitude: the F0 drop is largest at $1.5$B and $14$B and smallest at
$0.5$B and $7$B-base (per-size numbers in
Table~\ref{tab:per_model_frame}). Instruction tuning at
$7$B \emph{increases} the drop ($0.461 \to 0.549$) in every one of
the five frames. Cross-family
magnitudes fall within the Qwen envelope (Phi-3.5-mini $0.523$;
OLMo-2 $0.434$/$0.426$; SmolLM2 $0.327$/$0.138$), so the
inverted-U appears at every scale we test, but its magnitude is not
a clean function of size. A within-family depth-vs-width comparison
in Appendix~\ref{app:archcontrast} does not yield a clean account of
the SmolLM2-1.7B exception (\S\ref{sec:dense}).

\begin{table*}[!t]
  \centering
  \small
  \setlength{\tabcolsep}{6pt}
  \begin{tabular}{l c c c c c l}
    \toprule
    \textbf{Model} & \textbf{Obj.} & \textbf{Peak $N$} & \textbf{Peak $P$} & \textbf{$P(N{=}30)$} & \textbf{Drop} & \textbf{95\,\% CI of drop} \\
    \midrule
    Qwen2.5-14B~\citep{qwen2025qwen25} & CLM &    3 & 0.223 & 0.065 & 0.733 & $[0.706,\,0.761]$ \\
    Qwen2.5-1.5B~\citep{qwen2025qwen25} & CLM &    5 & 0.125 & 0.062 & 0.693 & $[0.654,\,0.733]$ \\
    Qwen2.5-7B-Instruct~\citep{qwen2025qwen25} & CLM &   10 & 0.209 & 0.155 & 0.549 & $[0.507,\,0.592]$ \\
    Phi-3.5-mini~\citep{abdin2024phi3} & CLM &    1 & 0.001 & 0.000 & 0.523 & $[0.485,\,0.558]$ \\
    ModernBERT-base~\citep{warner2024modernbert} & MLM &    3 & 0.668 & 0.372 & 0.516 & $[0.481,\,0.549]$ \\
    Qwen2.5-7B~\citep{qwen2025qwen25} & CLM &    5 & 0.134 & 0.106 & 0.461 & $[0.422,\,0.502]$ \\
    Qwen2.5-3B~\citep{qwen2025qwen25} & CLM &    3 & 0.109 & 0.040 & 0.434 & $[0.397,\,0.477]$ \\
    OLMo-2-1B~\citep{olmo2025olmo2} & CLM &   20 & 0.039 & 0.033 & 0.434 & $[0.387,\,0.472]$ \\
    OLMo-2-7B~\citep{olmo2025olmo2} & CLM &   10 & 0.058 & 0.042 & 0.426 & $[0.385,\,0.468]$ \\
    ModernBERT-large~\citep{warner2024modernbert} & MLM &    1 & 0.400 & 0.260 & 0.423 & $[0.388,\,0.462]$ \\
    Qwen2.5-0.5B~\citep{qwen2025qwen25} & CLM &    3 & 0.187 & 0.132 & 0.329 & $[0.294,\,0.363]$ \\
    SmolLM2-360M~\citep{allal2025smollm2} & CLM &   20 & 0.113 & 0.100 & 0.327 & $[0.291,\,0.367]$ \\
    SmolLM2-1.7B~\citep{allal2025smollm2} & CLM &   30 & 0.101 & 0.101 & 0.138 & $[0.114,\,0.164]$ \\
    \bottomrule
  \end{tabular}
  \caption{Displaced-probe F0 summary across the open-access panel
    (per-model target sets of $182$--$258$ single-token-valid words).
    \textit{Peak $N$}, \textit{Peak $P$} from the word-averaged mean
    curve; $P(N{=}30)$ the same curve at $N{=}30$. \textit{Drop} is
    the per-word mean of
    $(\text{peak}_w - P(N{=}30)_w)/\text{peak}_w$; $95\,\%$ CI
    bootstrapped over words ($500$ resamples, seed $42$). Every model
    has positive drop with CI excluding zero; SmolLM2-1.7B weakest,
    Qwen2.5-14B strongest.}
  \label{tab:model_results}
\end{table*}

\subsection{Truncation contrast}
\label{sec:truncation}

If the displaced curve really turns over before $N{=}30$, then
truncating the prefix back toward each model's per-word peak ought
to raise $P(\text{target})$. The most direct version of the test is
peak-relative: cut to each model's own peak $N$. We run the test
on the four-model probe subset used throughout for ablation and
mechanism (ModernBERT-large, Qwen2.5-1.5B, Qwen2.5-7B, OLMo-2-7B).
As a check on robustness we also run a fixed cut to $N{=}3$, which
serves as a stand-in for the peak in three of these four whose
displaced peak in fact sits at or near $N{=}3$ (ModernBERT-large at
$N{=}1$, Qwen2.5-1.5B and Qwen2.5-7B at $N{=}5$).

\paragraph{Peak-relative cut.} The per-word contrast
$P(\text{target}\mid N{=}\text{peak}) - P(\text{target}\mid N{=}30)$
is positive on all four probe models, with bootstrap intervals
clearing zero: $+0.141$ $[+0.114, +0.169]$ on ModernBERT-large
($N{=}30 \to 1$), $+0.063$ $[+0.050, +0.076]$ on Qwen2.5-1.5B
($N{=}30 \to 5$), $+0.028$ $[+0.011, +0.047]$ on Qwen2.5-7B
($N{=}30 \to 5$), and $+0.016$ $[+0.010, +0.023]$ on OLMo-2-7B
($N{=}30 \to 10$). This is not a new experiment so much as a sharper
read of the same per-$N$ grid that produced
Table~\ref{tab:model_results}.

\paragraph{Fixed $N{=}3$ cut (sensitivity).} The same contrast at the
hard-coded $N{=}3$ point confirms two of four cases directly:
ModernBERT-large gains $+0.034$ $[0.016, 0.056]$ with the target's
rank improving from $35$ to $7$, and Qwen2.5-1.5B gains $+0.059$
$[0.046, 0.072]$ with rank $75 \to 7$. Qwen2.5-7B moves in the same
direction in point estimate ($+0.015$, rank $10 \to 4$) but its CI
$[-0.005, 0.035]$ marginally crosses zero. OLMo-2-7B's fixed-$N{=}3$
contrast is a real negative ($-0.015$ $[-0.022, -0.009]$) only because
its displaced peak sits at $N{=}10$, so the cut to $N{=}3$ overshoots.
The prediction the inverted-U makes is ``cut the prefix back near the
peak and $P(\text{target})$ goes up'', not ``$N{=}3$ is privileged'':
under the peak-relative reading OLMo-2-7B is not exceptional and the
panel pattern is uniform.

\subsection{Cross-lingual replication}
\label{sec:lang}

\begin{figure*}[!htbp]
  \centering
  \includegraphics[width=\textwidth]{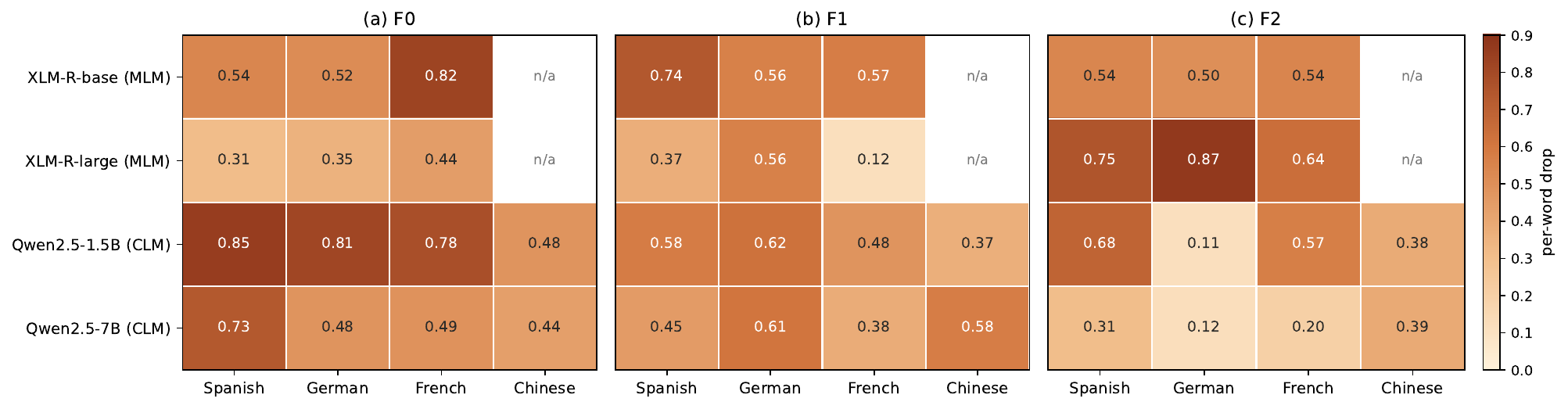}
  \caption{\textbf{Cross-lingual replication of the displaced
    inverted-U.} Per-word drop magnitude on the displaced probe for
    four open-access multilingual checkpoints across four target
    languages and three hand-translated frames. Each cell averages
    over $28$--$49$ single-token nouns; all 42 evaluated cells have
    bootstrap CIs that exclude zero. XLM-R Chinese cells (n/a)
    drop out because XLM-R's BPE retains no Chinese single-token
    candidates after filtering. Per-(model, language, frame)
    numbers are in Appendix~\ref{app:multiling},
    Table~\ref{tab:multilingual}.}
  \label{fig:crosslingual}
\end{figure*}

A reasonable concern after the English panel is that the displaced
inverted-U could be English-specific, or specific to WordPiece-like
tokenisation, or specific to the particular English frames we use.
The most straightforward way to rule those out is to rerun the displaced probe
unchanged in different languages and on tokenisers that do not look
like WordPiece. We probe four open-access multilingual checkpoints
(XLM-R-base, XLM-R-large \citep{conneau2020xlmr}, Qwen2.5-1.5B,
and Qwen2.5-7B \citep{qwen2025qwen25}) on Spanish, Chinese,
German, and French single-token noun lists, using three
hand-translated frames per language and $n_{w}{=}28$--$49$ per
(model, language) cell. XLM-R's BPE retains no Chinese single-token
candidates after filtering, so XLM-R Chinese drops out for both
base and large; what remains is fourteen (model, language)
combinations and $42$ (model, language, frame) cells in total
(Figure~\ref{fig:crosslingual}).

Every one of those $42$ cells shows a positive per-word drop with
its bootstrap interval clearing zero, and $30$ of $42$ also turn the
mean curve over before $N{=}30$ (the twelve that peak at $N{=}30$ keep
positive per-word drops; the breakdown is in
Appendix~\ref{app:multiling}). Magnitudes are large but spread, and
span similar ranges across the four languages. We
do not attempt a strong typological reading of the
language-by-language ordering: whether the magnitude differences
reflect tokeniser choices, morphological richness, or particular
frame translations cannot be resolved by this panel. What we can
conclude is that the displaced inverted-U is not an artefact
of any one of those three factors, since the shape recurs on two
multilingual MLMs and two multilingual CLMs, in four typologically
different target languages, every cell we measured.
Per-(model, language, frame) numbers are in
Appendix~\ref{app:multiling}, Table~\ref{tab:multilingual}.

\section{Exact Lexical Repetition Drives the Effect}
\label{sec:cause}

Section~\ref{sec:reversal} established the displaced inverted-U as a
panel-wide behavioural fact. We ask what the effect is and is not,
through three manipulations: a six-condition ablation that varies the
repeated block, a frame-pragmatics control that varies the readout
frame, and a pre-stated-answer test that gives the model a competing
retrieval route. The first two isolate the cause; the third bounds how
far the effect persists.

\begin{figure*}[!htbp]
  \centering
  \includegraphics[width=\textwidth]{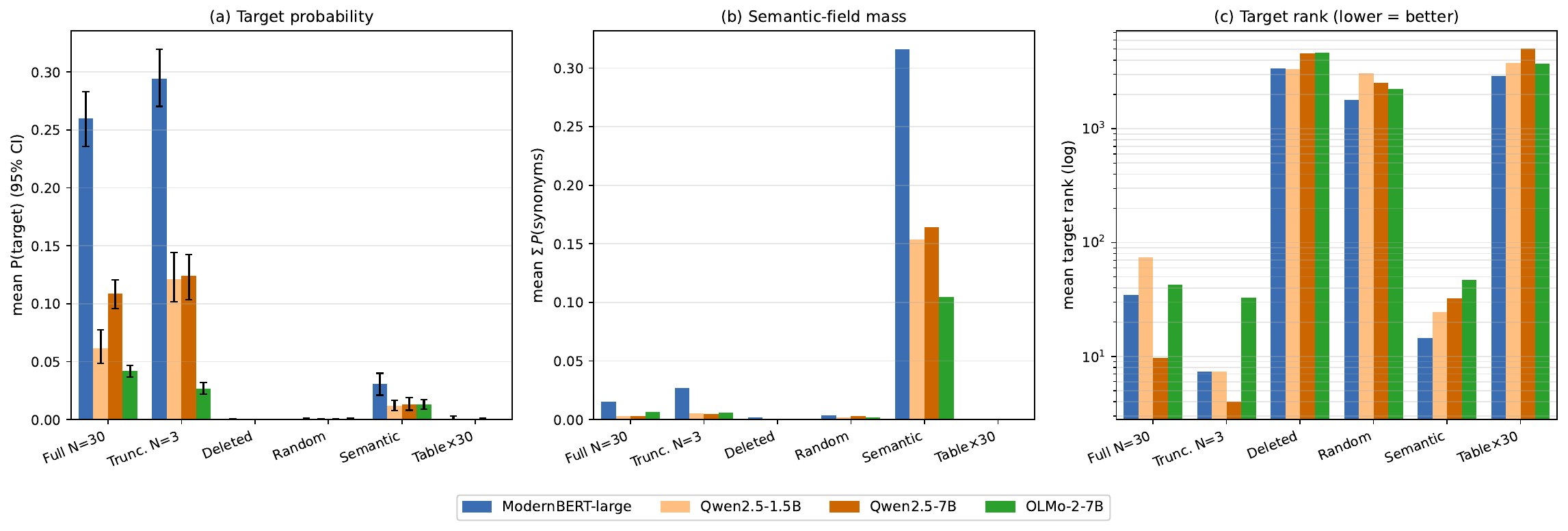}
  \caption{\textbf{Six-condition causal ablation, F0 displaced, four
    models} (shared legend below).
    (a) Mean $P(\text{target})$ per condition, $95\%$ bootstrap CIs.
    (b) $\Sigma P(\text{synonyms})$ per condition.
    (c) Mean target rank (log scale) per condition.
    Non-semantic conditions use the per-model single-token-valid set;
    the semantic-neighbour filler and $\Sigma P(\text{synonyms})$ use
    targets with curated semantic fields. Exact means in
    Appendix~\ref{app:ablation}, Table~\ref{tab:abl_full}.}
  \label{fig:ablation}
\end{figure*}

To pin down what in the prefix does the work, we swap the $N{=}30$
target block for five alternatives on the F0 displaced probe: a truncated
$N{=}3$ repeat of the same target, no block at all, a length-matched
random filler, a semantic-neighbour filler, and an unrelated-repeat
block (\codename{`table'}$\times 30$; constructions in
Appendix~\ref{app:ablation}).

Four of the five swaps eliminate the effect. Random filler,
deletion, and the unrelated-repeat block all push $P(\text{target})$
below $10^{-3}$ on all four ablation models. The interesting case is
the semantic-neighbour filler. It is the only condition that
\emph{activates} the target's semantic field: $\Sigma
P(\text{synonyms})$ climbs to $0.316$ in ModernBERT-large and to
$0.10$--$0.16$ in the three CLMs (Figure~\ref{fig:ablation}a,b). But
$P(\text{target})$ itself stays at $0.012$--$0.031$, well below the
full-repeat range of $0.042$--$0.260$. Semantic-field activation and
target-specific priming come apart cleanly. The dissociation is
order-of-magnitude: under full repetition, $P(\text{target})$
exceeds $\Sigma P(\text{synonyms})$ by $7$--$36\times$; under
semantic-neighbour fill, the ratio reverses, with
$\Sigma P(\text{synonyms})$ exceeding $P(\text{target})$ by
$8$--$13\times$.

Whatever the displaced
inverted-U is, it is not driven by sequence length, not by
repetition in general, and not by exposure to neighbours of the
target. It is exact lexical repetition of the target itself,
interacting with the displaced readout. That isolates the source
side of the displaced probe.

\begin{table*}[!t]
  \centering
  \small
  \setlength{\tabcolsep}{8pt}
  \begin{tabular}{l c l c}
    \toprule
    \textbf{Model} & \textbf{truncate $-$ full} & \textbf{95\,\% CI} & \textbf{Mean target rank: full $\to$ truncate} \\
    \midrule
    ModernBERT-large & $+0.034$ & $[0.016,\,0.056]$ & $35 \to 7$ \\
    Qwen2.5-1.5B & $+0.059$ & $[0.046,\,0.072]$ & $75 \to 7$ \\
    Qwen2.5-7B & $+0.015$ & $[-0.005,\,0.035]$ & $10 \to 4$ \\
    OLMo-2-7B & $-0.015$ & $[-0.022,\,-0.009]$ & $43 \to 33$ \\
    \bottomrule
  \end{tabular}
  \caption{Causal ablation: truncating $N{=}30$ to $N{=}3$ on F0
    displaced. Contrast is the per-word mean
    $P(\text{target} \mid N{=}3) - P(\text{target} \mid N{=}30)$;
    $95\,\%$ CI bootstrapped over per-model single-token-valid
    targets. Two of four models replicate with CI excluding zero
    (ModernBERT-large $\approx 5\times$ rank improvement,
    Qwen2.5-1.5B $\approx 11\times$); Qwen2.5-7B's point estimate
    is positive but CI includes zero. OLMo-2-7B's CI excludes zero
    in the opposite direction: its displaced peak sits at
    $N{=}10$, so $N{=}3$ lands before the peak.}
  \label{tab:ablation}
\end{table*}

\paragraph{Frame pragmatics.}
\label{sec:framepragm}
The displaced probe also carries a frame, and F0--F4 all introduce a
word, so the inverted-U might track the frame's pull toward a
\emph{new} word rather than non-additive aggregation. Fixing the
block and varying only the frame's pragmatic class on the four-model
subset rules this out: the per-word drop clears zero in all $16$
(model, class) cells; anaphoric frames that \emph{demand} the
repeated word still invert in every model, and novelty frames are the
largest-drop class in none (Appendix~\ref{app:framepragm}).

\paragraph{Persistence with a pre-stated answer.}
\label{sec:answerpresent}
A LAMA-style pilot supplies the target as an explicit answer at the
start of the prompt, a retrieval route the displaced probe withholds,
and then appends the repeated block before the readout. Three of four
mechanism models still retain a positive per-word drop with CI
excluding zero (Appendix~\ref{app:lama}). The magnitudes are
attenuated by $75$--$86\%$ from the main displaced range
($0.42$--$0.69$ becomes $0.058$--$0.136$), but the signature survives:
an alternative source of the answer does not eliminate the displaced
inverted-U.

\section{Internal Correlates}
\label{sec:mech}

The behavioural picture is established. The next question is what
inside the model moves alongside it. We run two attention probes and
a slot-state probe on four panel models chosen to span both
objectives and three scales: ModernBERT-large, Qwen2.5-1.5B,
Qwen2.5-7B, and OLMo-2-7B. The \emph{slot-state} probe reads the
target's probability off the logit head at the readout slot; the
attention probes measure how much attention that slot pays back to
the repeated block. Each runs on a word subset stratified by
concreteness and frequency: $10$ words for the attention probes, $23$
for the slot-state probe, and $50$ for the layer-wise breakdown
(Appendix~\ref{app:layers}). These probes are correlational
\citep{belinkov2022probing}: we record internal patterns that
co-move with the behavioural effect, and do not attempt to
adjudicate among mechanisms that produce the same co-movement.

Two terms recur below. We call per-target-token attention staying
\emph{above} its $1/N$ baseline \textit{sub-dilution}: the block
holds more attention than equal division across $N$ keys would
predict. We call per-token attention falling \emph{below} the
$1/N$ baseline \textit{super-dilution}: the block sheds attention
faster than equal division would predict.

\begin{figure*}[!htbp]
  \centering
  \includegraphics[width=\textwidth]{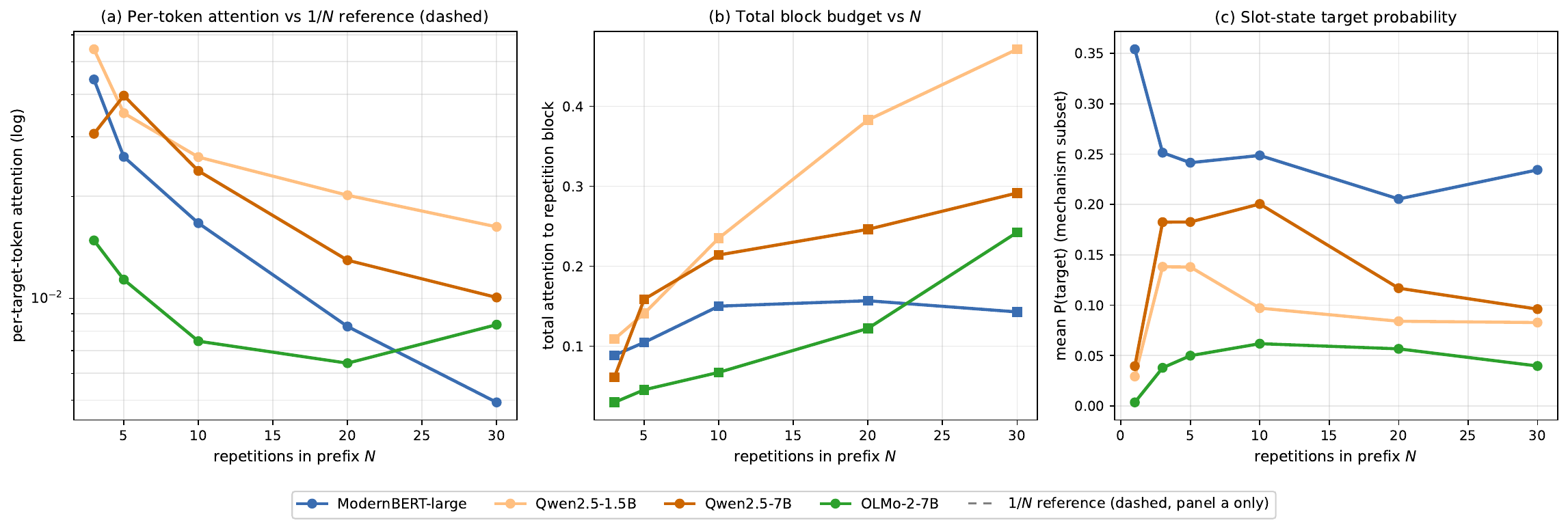}
  \caption{\textbf{Internal correlates on four panel models}
    (10/23-word mechanism subsets, F0 displaced; shared legend).
    (a) Per-target-token attention (solid, log) vs each model's
    $1/N$ reference (dashed, anchored at smallest $N$): attention
    drops with $N$ in every model. ModernBERT-large tracks its
    $1/N$ reference; the three CLMs sit above theirs (sub-dilution).
    (b) Total block attention: grows with $N$ for the three CLMs
    (sub-dilution); plateaus for ModernBERT-large.
    (c) Slot-state target probability ($23$-word subset): in the
    three CLMs, slot-state target probability tracks the behavioural
    inverted-U; ModernBERT-large peaks earliest and decays unevenly.}
  \label{fig:mechanism}
\end{figure*}

\paragraph{Total block attention grows with $N$.}
Per-target-token attention falls with $N$ in all four mechanism
models (Figure~\ref{fig:mechanism}a), as budget conservation would
predict if a fixed budget were divided over $N$ keys. The total
attention the model assigns to the repeated block, however, does not
stay fixed. In three of four models the block budget grows with $N$
(Figure~\ref{fig:mechanism}b): Qwen2.5-1.5B from $\approx 0.11$ at
$N{=}3$ to $\approx 0.47$ at $N{=}30$; Qwen2.5-7B from $0.06$ to
$0.29$; OLMo-2-7B from $0.03$ to $0.24$. ModernBERT-large is the
exception, with a roughly flat block budget ($0.09$ to $0.14$;
per-$N$ table in Appendix~\ref{app:layers}). The per-token fall in
the three CLMs is therefore sub-dilution: per-token attention drops
because the budget is divided across more tokens, but the budget
itself grows.

A layer-wise breakdown of the same per-target-token attention
(Appendix~\ref{app:layers}, Figure~\ref{fig:layerattn}) confirms
that sub-dilution is a stack-wide property rather than a last-layer
artefact: the three CLMs sit above their per-layer $1/N$ reference
at most depths, while ModernBERT-large super-dilutes in its upper
layers, consistent with its roughly flat last-layer block budget.
The displaced inverted-U's attention correlate dissociates by
training objective in this subset: CLMs allocate more total
attention to the repeated block as $N$ grows; the MLM does not.

\paragraph{Slot-state target probability tracks the behavioural
inverted-U.}
On the $23$-word mechanism subset, the model's logit-head
probability for the target shows a clean inverted-U on the three
CLMs: Qwen2.5-1.5B peaks at $N{=}3$
($0.138$), Qwen2.5-7B peaks around $N{=}5$--$10$ ($0.18$--$0.20$),
and OLMo-2-7B peaks at $N{=}10$ ($0.062$); all decay through $N{=}30$
(Figure~\ref{fig:mechanism}c). ModernBERT-large peaks earliest at
$N{=}1$ ($0.354$) and decays unevenly.

\section{Discussion}
\label{sec:discussion}

The methodological lesson sits in a single contrast. Hold the
repeated-target prefix fixed, move the prediction slot off the block,
and the curve no longer climbs: it peaks early and falls back. The
same prefix produces priming on the adjacent slot in every baseline
we test, so the displaced decline cannot be reduced to long-context
capacity loss. The consequence for the probing studies surveyed in
\S\ref{sec:intro} is that the slot a study scores is not a neutral
measurement choice: a manipulation that reads as monotone priming at
one position can invert one clause later, and the cross-lingual
replication argues this is not an English- or tokeniser-specific
concern.

The substantive claim is that repetition helps when the readout slot
extends the repeated surface, and hurts when the target must be
retrieved from a fresh clause. The six-condition ablation traces the
active variable to exact lexical repetition rather than prefix length
or semantic-neighbourhood activation.

The diagnostic is weakest on SmolLM2-1.7B, whose mean displaced curve
does not visibly invert, though its per-word drop stays positive and
the headline statistic holds across the full $13$-model panel. The
within-family comparison with SmolLM2-360M cuts against parameter
count: the
smaller, deeper model shows a clearer inversion than the larger,
shallower one. The same direction shows up on BERT-tiny/mini/small
(Appendix~\ref{app:archcontrast}): the $2$-layer BERT-tiny shows no
inverted-U, while $4$-layer BERT-mini and BERT-small do (drops
$0.521$ and $0.478$). Depth, not parameter
count, is the more relevant axis in these comparisons.

Post-training does not fix the effect either. Instruction tuning at
$7$B does not attenuate the displaced drop but \emph{increases} it
(Qwen2.5-7B base $0.461 \to$ Qwen2.5-7B-Instruct $0.549$ on F0, same
direction in all five frames).

The internal correlate points the same way. The per-target-token
attention drop is what budget conservation predicts on its own; the
growth in total block budget is not, and it splits by training
objective rather than scale (\S\ref{sec:mech}), consistent with
autoregressive training doing something with repeated keys that an
MLM does not. Whether the underlying mechanism is softmax cooperation
between repeated keys, attention-sink dynamics
\citep{xiao2024streaming, gu2024when} under a displaced-readout setup,
or autoregressive-specific recency priors, our correlational probes
cannot distinguish, and a single MLM is not enough to tell the
candidates apart.

\section*{Limitations}

Every target in our English panel passes a single-token filter in
every tested tokeniser family, which excludes many naturalistic
content words. The obvious extension (multi-token targets scored
under probability-product or first-piece heuristics) is not a
free generalisation: the headline statistic and the ablation
conditions both assume a single readout position. The non-additive
aggregation claim is bounded accordingly to single-token cloze and
next-token prediction under the displaced-readout frame structure
of \S\ref{sec:setup}; whether the same shape appears in free-form
generation, multi-hop retrieval, or under multi-token scoring is a
separate question.

The two probes are unevenly covered. The adjacent baseline in
Figure~\ref{fig:main_curves}a runs on only three models (one MLM, one
small CLM, one larger CLM) rather than the full $13$; the dissociation
requires the contrast to hold at all, not model-by-model, but
extending the adjacent probe to the full panel would strengthen the
argument.

The cross-lingual frames F0--F2 are hand-translated to
idiomatic equivalents, without calibration for matched syntactic
predictability, so frame-by-frame magnitudes across languages may
partly reflect translation choices; the panel-level claim ($42$ of
$42$ cells positive) does not depend on those comparisons.

SmolLM2-1.7B is an exception to the secondary descriptive
observation (mean-curve inversion before $N{=}30$) but not to the
headline statistic (positive per-word drop with CI excluding zero).
We do not have a clean account of why this particular checkpoint's
mean curve fails to invert.

Per-word drops are computed only on
words with $\text{peak}_w > 0$; Phi-3.5-mini's absolute
$P(\text{target})$ is much lower than the rest of the panel
($\text{peak}_P \approx 0.001$ at the mean-curve level), and its
$13$-of-$13$ contribution rests on relative collapse across this
low-probability regime rather than on a meaningful absolute decay.

The frame-pragmatics ablation is English-only on the four-model
subset, and its novelty frames are a weaker manipulation check than
frames pre-validated to shift continuations toward novel words.

The internal-correlate evidence is correlational. Per-target-token
attention drops, the slot-state inverted-U turns over, and the
layer-wise picture is structured rather than diffuse; all co-move
with the behaviour, but co-movement is not intervention. None of our
probes manipulates the attention pattern or the displaced slot
directly; calibrated interventions that mask the repeated block's
attention or perturb the slot-state representation remain open.

\bibliography{references}

\appendix

\section{Word List, Tokeniser Filtering, and Templates}
\label{app:wordlist}

\paragraph{Construction.}
We hand-curate $261$ candidate English nouns and noun-like concepts.
Each candidate is annotated with its Zipf log-frequency
\citep{speer2022wordfreq} and a semantic class. The list is balanced
across four categorical cells (cross of concrete vs abstract
reference, and high vs low frequency, with the high/low boundary at
Zipf $5.0$) and across twelve semantic classes.

\paragraph{Per-tokeniser filtering.}
Tokenisers fragment many candidates differently, so we
construct a per-family target set by retaining candidates that
encode to exactly one piece under that family's tokeniser. For each
candidate we try both the leading-space form (BPE / SentencePiece
convention) and the no-space form (BERT WordPiece convention) and
accept the candidate if either yields a single token. This produces
target sets of $182$ (Phi-3.5-mini) to $258$ (SmolLM2) words. The
shared core (the intersection across every panel tokeniser) is
$\approx 168$ words and is the basis for the cross-model OLS
regression of \S\ref{sec:stat}.

\paragraph{Frame templates.}
\textit{Adjacent} probes (no displaced frame):
\begin{itemize}\itemsep0pt
\item MLM: \codename{<target $\times$ N> [MASK] .}
\item CLM: \codename{<target $\times$ N>} (next-token slot).
\end{itemize}
\textit{Displaced} probes:
\begin{itemize}\itemsep0pt
\item MLM: \codename{<target $\times$ N> . <frame> [MASK] .}
\item CLM: \codename{<target $\times$ N> . <frame>} (next-token slot).
\end{itemize}
\textit{Frames F0--F4 (English):}
\begin{itemize}\itemsep0pt
\item F0: \emph{The word I keep thinking about is~\dots}
\item F1: \emph{The topic was~\dots}
\item F2: \emph{I want to mention~\dots}
\item F3: \emph{Let me tell you about~\dots}
\item F4: \emph{The most important thing is~\dots}
\end{itemize}
Ablation block templates that replace
\codename{<target $\times$ 30>} in the F0 displaced probe are
defined in Appendix~\ref{app:ablation}.

\section{Statistical Analysis: Full Tables}
\label{app:stats}

\paragraph{Quadratic regression: coefficients.}
The primary OLS quadratic on shared-core $N \geq 1$ cells
($n = 70{,}590$) yields the parameter table below; cluster-robust
SEs (clustered by target word).
Reference levels: model ModernBERT-base, frame F0, category
abstract-high. $R^{2} = 0.550$.

\begin{table}[!htbp]
  \centering
  \footnotesize
  \setlength{\tabcolsep}{6pt}
  \begin{tabular}{l r}
    \toprule
    \textbf{Term} & \textbf{Coefficient} \\
    \midrule
    Intercept                        & $-0.725$ \\
    $N$                              & $+0.2118$ \\
    $N^{2}$                          & $-0.00610$ \\
    Zipf log-frequency               & $-0.191$ \\
    \midrule
    model: ModernBERT-large          & $-0.43$ \\
    model: Qwen2.5-0.5B              & $-3.10$ \\
    model: Qwen2.5-1.5B              & $-3.51$ \\
    model: Qwen2.5-3B                & $-2.84$ \\
    model: Qwen2.5-7B                & $-1.95$ \\
    model: Qwen2.5-7B-Instruct       & $-2.10$ \\
    model: Qwen2.5-14B               & $-2.05$ \\
    model: SmolLM2-360M              & $-3.45$ \\
    model: SmolLM2-1.7B              & $-2.82$ \\
    model: Phi-3.5-mini              & $-7.34$ \\
    model: OLMo-2-1B                 & $-3.63$ \\
    model: OLMo-2-7B                 & $-2.90$ \\
    \midrule
    frame: F1                        & $-1.77$ \\
    frame: F2                        & $-2.29$ \\
    frame: F3                        & $-1.12$ \\
    frame: F4                        & $-2.27$ \\
    \midrule
    category: abstract-low           & $-0.63$ \\
    category: concrete-high          & $-0.27$ \\
    category: concrete-low           & $-0.84$ \\
    \bottomrule
  \end{tabular}
  \caption{OLS quadratic regression coefficients on
    $\log P(\text{target})$ over the shared-core panel, rounded for
    display. Both $N$ and $N^{2}$ signs match the inverted-U
    prediction; the largest fixed effects are by model.}
  \label{tab:ols_full}
\end{table}

\paragraph{Pattern classification thresholds.}
Each per-word $N$-curve is classified by deterministic rules. Let
$\text{peak}_w = \max_{N \in \{1,3,5,10,20,30\}} P(\text{target} \mid w, N)$
and let $N^{*}_{w}$ denote the argmax. A (model, frame, word) cell
is classified as
\begin{itemize}\itemsep0pt
\item \textit{satiation-like} if $N^{*}_{w} \in \{1,3,5,10\}$
  \emph{and} $(\text{peak}_w - P(N{=}30 \mid w))/\text{peak}_w > 0.30$;
\item \textit{monotonic-priming} if the curve is non-decreasing in
  $N$ \emph{and} $P(N{=}30 \mid w) > 1.05 \cdot P(N{=}1 \mid w)$;
\item \textit{plateau} if the curve is approximately monotonic with
  $P(N{=}30 \mid w) \geq 0.7 \cdot \text{peak}_w$;
\item \textit{mixed} otherwise.
\end{itemize}
Across the $16{,}230$ (model, frame, word) cells, $8{,}964$
($55\%$) are \textit{satiation-like}, $5{,}129$ ($32\%$) are
\textit{plateau}, $1{,}089$ ($7\%$) are \textit{monotonic-priming},
and $1{,}048$ ($6\%$) are \textit{mixed}. Per-(model, frame) shares
appear in the rightmost column of
Table~\ref{tab:per_model_frame}.

\paragraph{Frequency confound regression.}
Table~\ref{tab:freq_reg} reports the targeted regression
$\text{drop}_w \sim \text{Zipf} + \text{category} + \text{model} +
\text{frame}$ on $n{=}16{,}230$ word-frame cells with cluster-robust
SEs by word. The full regression has $R^{2}{=}0.191$; Zipf alone has
$R^{2}{=}0.003$ and Zipf+category together $R^{2}{=}0.004$, while
model+frame fixed effects alone account for $R^{2}{=}0.186$, so the
fixed effects explain $\approx 60\times$ more variance than Zipf
alone. The Zipf coefficient is small but reliably negative under
clustering ($-0.036$, $t{=}-3.5$, $p < 10^{-3}$), so high-frequency
words decay slightly less than low-frequency ones; the category
coefficients lose individual significance under clustering
(concrete-low: $t = -1.56$, $p = 0.12$; others $|t| < 1.5$), so we
do not interpret per-category differences in decay magnitude.
\textbf{Frequency modulates the magnitude of the decay but does not
explain it away.}

\begin{table}[!htbp]
  \centering
  \footnotesize
  \setlength{\tabcolsep}{6pt}
  \begin{tabular}{l r}
    \toprule
    \textbf{Term} & \textbf{Coefficient} \\
    \midrule
    Zipf log-frequency               & $-0.036$ \\
    category: abstract-low           & $+0.015$ \\
    category: concrete-high          & $-0.012$ \\
    category: concrete-low           & $-0.029$ \\
    \midrule
    \multicolumn{2}{l}{$R^{2}$ (full, with model+frame): $0.191$} \\
    \multicolumn{2}{l}{$R^{2}$ (Zipf only): $0.003$} \\
    \multicolumn{2}{l}{$R^{2}$ (Zipf $+$ category only): $0.004$} \\
    \multicolumn{2}{l}{$R^{2}$ (model $+$ frame only): $0.186$} \\
    \multicolumn{2}{l}{$n = 16{,}230$ word-frame cells} \\
    \bottomrule
  \end{tabular}
  \caption{Frequency-confound regression $\text{drop}_w \sim
    \text{Zipf} + \text{category} + \text{model} + \text{frame}$;
    cluster-robust SEs (clustered by target word). Zipf has the
    right-sign coefficient but explains an order of magnitude less
    variance than the (model, frame) fixed effects.}
  \label{tab:freq_reg}
\end{table}

\paragraph{Per-(model, frame) summary: complete table.}
Table~\ref{tab:per_model_frame} reports peak $N$, peak $P$,
$P(N{=}30)$, drop, bootstrap CI, and \textit{satiation-like} share
for every (model, frame) cell ($65$ rows: $13$ models $\times$ $5$
frames). SmolLM2-1.7B is the only row block
whose mean curve does not invert on any frame; its per-word drops
remain positive but the \textit{satiation-like} share never exceeds
$0.18$.

\begin{table*}[!htbp]
  \centering
  \footnotesize
  \setlength{\tabcolsep}{4pt}
  \renewcommand{\arraystretch}{0.95}
  \begin{tabular}{l c c c c c c c}
    \toprule
    \textbf{Model} & \textbf{Frame} & \textbf{Peak} $N$ & \textbf{Peak} $P$ & \textbf{$P(N{=}30)$} & \textbf{Drop} & \textbf{95\% CI of drop} & \textbf{Sat.\,share} \\
    \midrule
    ModernBERT-base    & F0 &    3 & 0.668 & 0.372 & 0.516 & $[0.481,\,0.549]$ & 0.74 \\
                       & F1 &    5 & 0.292 & 0.107 & 0.728 & $[0.696,\,0.761]$ & 0.90 \\
                       & F2 &    3 & 0.132 & 0.044 & 0.696 & $[0.665,\,0.728]$ & 0.91 \\
                       & F3 &    3 & 0.329 & 0.091 & 0.787 & $[0.759,\,0.812]$ & 0.93 \\
                       & F4 &    3 & 0.119 & 0.034 & 0.832 & $[0.805,\,0.855]$ & 0.96 \\
    ModernBERT-large   & F0 &    1 & 0.400 & 0.260 & 0.423 & $[0.388,\,0.462]$ & 0.65 \\
                       & F1 &   10 & 0.210 & 0.141 & 0.455 & $[0.415,\,0.495]$ & 0.56 \\
                       & F2 &   10 & 0.062 & 0.050 & 0.382 & $[0.348,\,0.422]$ & 0.46 \\
                       & F3 &   10 & 0.146 & 0.086 & 0.551 & $[0.514,\,0.585]$ & 0.74 \\
                       & F4 &    1 & 0.160 & 0.084 & 0.570 & $[0.534,\,0.604]$ & 0.76 \\
    Qwen2.5-0.5B       & F0 &    3 & 0.187 & 0.132 & 0.329 & $[0.294,\,0.363]$ & 0.53 \\
                       & F1 &    3 & 0.012 & 0.008 & 0.398 & $[0.350,\,0.444]$ & 0.51 \\
                       & F2 &    3 & 0.014 & 0.006 & 0.589 & $[0.553,\,0.621]$ & 0.80 \\
                       & F3 &    3 & 0.015 & 0.014 & 0.256 & $[0.218,\,0.292]$ & 0.36 \\
                       & F4 &   30 & 0.005 & 0.005 & 0.247 & $[0.214,\,0.280]$ & 0.33 \\
    Qwen2.5-1.5B       & F0 &    5 & 0.125 & 0.062 & 0.693 & $[0.654,\,0.733]$ & 0.81 \\
                       & F1 &    5 & 0.023 & 0.014 & 0.684 & $[0.639,\,0.730]$ & 0.71 \\
                       & F2 &    3 & 0.010 & 0.002 & 0.825 & $[0.797,\,0.856]$ & 0.92 \\
                       & F3 &    3 & 0.039 & 0.008 & 0.856 & $[0.829,\,0.884]$ & 0.94 \\
                       & F4 &    5 & 0.008 & 0.004 & 0.629 & $[0.585,\,0.672]$ & 0.72 \\
    Qwen2.5-3B         & F0 &    3 & 0.109 & 0.040 & 0.434 & $[0.397,\,0.477]$ & 0.60 \\
                       & F1 &   30 & 0.014 & 0.014 & 0.418 & $[0.372,\,0.462]$ & 0.52 \\
                       & F2 &    3 & 0.006 & 0.004 & 0.481 & $[0.444,\,0.516]$ & 0.59 \\
                       & F3 &    3 & 0.054 & 0.026 & 0.551 & $[0.515,\,0.587]$ & 0.71 \\
                       & F4 &    3 & 0.008 & 0.003 & 0.436 & $[0.401,\,0.471]$ & 0.60 \\
    Qwen2.5-7B         & F0 &    5 & 0.134 & 0.106 & 0.461 & $[0.422,\,0.502]$ & 0.52 \\
                       & F1 &   30 & 0.029 & 0.029 & 0.282 & $[0.247,\,0.320]$ & 0.33 \\
                       & F2 &   30 & 0.018 & 0.018 & 0.393 & $[0.354,\,0.434]$ & 0.51 \\
                       & F3 &   30 & 0.045 & 0.045 & 0.283 & $[0.250,\,0.321]$ & 0.37 \\
                       & F4 &   30 & 0.115 & 0.115 & 0.209 & $[0.171,\,0.249]$ & 0.18 \\
    Qwen2.5-7B-Instruct & F0 &  10 & 0.209 & 0.155 & 0.549 & $[0.507,\,0.592]$ & 0.55 \\
                       & F1 &    5 & 0.027 & 0.024 & 0.512 & $[0.465,\,0.553]$ & 0.58 \\
                       & F2 &   10 & 0.014 & 0.011 & 0.494 & $[0.452,\,0.535]$ & 0.58 \\
                       & F3 &   30 & 0.088 & 0.088 & 0.376 & $[0.343,\,0.419]$ & 0.44 \\
                       & F4 &   30 & 0.097 & 0.097 & 0.281 & $[0.236,\,0.327]$ & 0.29 \\
    Qwen2.5-14B        & F0 &    3 & 0.223 & 0.065 & 0.733 & $[0.706,\,0.761]$ & 0.93 \\
                       & F1 &    5 & 0.028 & 0.015 & 0.499 & $[0.464,\,0.530]$ & 0.71 \\
                       & F2 &    5 & 0.018 & 0.010 & 0.479 & $[0.447,\,0.512]$ & 0.69 \\
                       & F3 &    3 & 0.083 & 0.031 & 0.674 & $[0.645,\,0.704]$ & 0.89 \\
                       & F4 &    5 & 0.021 & 0.006 & 0.545 & $[0.511,\,0.577]$ & 0.78 \\
    SmolLM2-360M       & F0 &   20 & 0.113 & 0.100 & 0.327 & $[0.291,\,0.367]$ & 0.29 \\
                       & F1 &   20 & 0.006 & 0.006 & 0.267 & $[0.233,\,0.301]$ & 0.27 \\
                       & F2 &   20 & 0.009 & 0.008 & 0.365 & $[0.326,\,0.407]$ & 0.42 \\
                       & F3 &   20 & 0.032 & 0.030 & 0.262 & $[0.223,\,0.300]$ & 0.28 \\
                       & F4 &   20 & 0.003 & 0.003 & 0.328 & $[0.294,\,0.365]$ & 0.41 \\
    SmolLM2-1.7B       & F0 &   30 & 0.101 & 0.101 & 0.138 & $[0.114,\,0.164]$ & 0.05 \\
                       & F1 &   30 & 0.032 & 0.032 & 0.255 & $[0.221,\,0.290]$ & 0.14 \\
                       & F2 &   30 & 0.019 & 0.019 & 0.206 & $[0.171,\,0.240]$ & 0.18 \\
                       & F3 &   30 & 0.040 & 0.040 & 0.191 & $[0.161,\,0.224]$ & 0.09 \\
                       & F4 &   30 & 0.026 & 0.026 & 0.213 & $[0.181,\,0.247]$ & 0.05 \\
    Phi-3.5-mini       & F0 &    1 & 0.001 & 0.000 & 0.523 & $[0.485,\,0.558]$ & 0.75 \\
                       & F1 &    1 & 0.000 & 0.000 & 0.459 & $[0.421,\,0.497]$ & 0.69 \\
                       & F2 &    1 & 0.000 & 0.000 & 0.608 & $[0.574,\,0.642]$ & 0.92 \\
                       & F3 &   10 & 0.000 & 0.000 & 0.333 & $[0.295,\,0.373]$ & 0.50 \\
                       & F4 &    1 & 0.000 & 0.000 & 0.564 & $[0.528,\,0.602]$ & 0.84 \\
    OLMo-2-1B          & F0 &   20 & 0.039 & 0.033 & 0.434 & $[0.387,\,0.472]$ & 0.48 \\
                       & F1 &   20 & 0.013 & 0.013 & 0.500 & $[0.451,\,0.545]$ & 0.55 \\
                       & F2 &   30 & 0.005 & 0.005 & 0.291 & $[0.244,\,0.335]$ & 0.29 \\
                       & F3 &   20 & 0.024 & 0.022 & 0.301 & $[0.261,\,0.343]$ & 0.31 \\
                       & F4 &   30 & 0.010 & 0.010 & 0.225 & $[0.187,\,0.266]$ & 0.24 \\
    OLMo-2-7B          & F0 &   10 & 0.058 & 0.042 & 0.426 & $[0.385,\,0.468]$ & 0.43 \\
                       & F1 &   10 & 0.022 & 0.015 & 0.412 & $[0.372,\,0.454]$ & 0.50 \\
                       & F2 &   10 & 0.008 & 0.005 & 0.474 & $[0.433,\,0.515]$ & 0.56 \\
                       & F3 &   10 & 0.041 & 0.022 & 0.537 & $[0.495,\,0.579]$ & 0.67 \\
                       & F4 &   10 & 0.011 & 0.005 & 0.565 & $[0.529,\,0.601]$ & 0.72 \\
    \bottomrule
  \end{tabular}
  \caption{Per-(model, frame) summary
    ($n_{\mathrm{words}}$ ranges from $182$ to $258$). \textit{Peak
    $N$} and \textit{Peak $P$} are taken from the word-averaged mean
    curve. \textit{Drop} is the per-word mean of $(\text{peak}_w -
    P(N{=}30)_w)/\text{peak}_w$; the $95\%$ CI is bootstrapped over
    the per-model single-token-valid target words.
    \textit{Sat.\,share} is the fraction of (word) curves classified
    as \textit{satiation-like}. Model name shown on the F0 row of
    each block; blank below.}
  \label{tab:per_model_frame}
\end{table*}

\section{Within-Family Architecture Contrast}
\label{app:archcontrast}

The two SmolLM2 checkpoints provide a small natural
depth-vs-capacity dissociation. SmolLM2-360M has $32$ transformer
blocks at hidden size $960$ ($15$ attention heads, $2560$
intermediate dimension) and an F0 drop of $0.327$. SmolLM2-1.7B has
fewer blocks ($24$) at a wider hidden size ($2048$, $32$ attention
heads, $8192$ intermediate) and an F0 drop of $0.138$. The larger
checkpoint has more parameters but is shallower, and shows the weaker
effect; the smaller, deeper checkpoint shows the stronger effect.
Within the SmolLM2 family, depth therefore correlates with
displaced-inverted-U magnitude in the opposite direction from raw
parameter count. We treat this as suggestive rather than conclusive
(a single family with one depth-vs-width contrast is not a
controlled scaling law).

\paragraph{BERT-tiny / mini / small.}
The same direction shows up on the older
\citet{turc2019well} BERT family, which we ran on the F0 displaced
probe as an auxiliary check. BERT-tiny ($2$ layers, hidden $128$,
$4.4$M params) shows no inverted-U: F0 peak $N{=}30$, per-word drop
$0.001$ $[0.000, 0.003]$. BERT-mini ($4$ layers, hidden $256$,
$11$M) shows a clear inverted-U: F0 peak $N{=}3$, per-word drop
$0.521$ $[0.471, 0.565]$. BERT-small ($4$ layers, hidden $512$,
$29$M) also shows a per-word inversion with CI clearing zero
($0.478$ $[0.426, 0.528]$), though its mean curve peaks at $N{=}30$.
The transition between absent and present sits at the depth jump
from $2$ to $4$ layers; doubling hidden size at fixed depth
(mini~$\to$~small) leaves the per-word drop essentially unchanged.
This is consistent with the SmolLM2 within-family direction.

\section{Causal Ablation Construction}
\label{app:ablation}

\paragraph{Condition definitions.}
On the F0 displaced probe, each input replaces the
$30$-target-repeat block with one of six conditions. The
non-semantic conditions are run over the per-model
single-token-valid set; the semantic-neighbour filler and the
semantic-field mass are restricted to targets with curated semantic
dictionaries. The full $N{=}30$ repeat (code:
\codename{full\_repeat\_N30}) keeps thirty exact copies of the
target. The truncated $N{=}3$ repeat (code:
\codename{truncate\_to\_N3}) keeps only the first three target
copies. The deleted block (code: \codename{delete\_repeat\_block})
is empty. The length-matched random filler (code:
\codename{random\_filler\_same\_length}) is thirty random nouns
sampled from a fixed neutral pool. The semantic-neighbour filler
(code: \codename{unique\_semantic\_filler}) is thirty semantic
neighbours sampled from the target's curated semantic field,
excluding the exact target. The unrelated-repeat block (code:
\codename{repeat\_unrelated\_N30}) is thirty copies of
\codename{`table'}.

\paragraph{Per-(model, condition) summary.}
Table~\ref{tab:abl_full} reports means over the per-model target
set for each condition. The full $N{=}30$ repeat and the truncated
$N{=}3$ repeat are the only conditions with non-trivial
$P(\text{target})$; the semantic-neighbour filler is the only
condition that raises $\Sigma P(\text{synonyms})$ substantially,
illustrating the separation between target-specific lexical priming
and semantic-neighbourhood activation.

\begin{table*}[!htbp]
  \centering
  \footnotesize
  \setlength{\tabcolsep}{4pt}
  \begin{tabular}{l l r r}
    \toprule
    \textbf{Model} & \textbf{Condition} & $P(\text{target})$ & $\Sigma P(\text{syn.})$ \\
    \midrule
    ModernBERT-large & Full $N{=}30$ repeat        & 0.260 & 0.015 \\
    ModernBERT-large & Truncated $N{=}3$ repeat    & 0.294 & 0.027 \\
    ModernBERT-large & Deleted block               & 0.001 & 0.002 \\
    ModernBERT-large & Random filler               & 0.001 & 0.004 \\
    ModernBERT-large & Semantic-neighbour filler   & 0.031 & \textbf{0.316} \\
    ModernBERT-large & Unrelated-repeat block      & 0.001 & 0.001 \\
    \midrule
    Qwen2.5-1.5B     & Full $N{=}30$ repeat        & 0.062 & 0.003 \\
    Qwen2.5-1.5B     & Truncated $N{=}3$ repeat    & 0.121 & 0.005 \\
    Qwen2.5-1.5B     & Deleted block               & 0.000 & 0.001 \\
    Qwen2.5-1.5B     & Random filler               & 0.000 & 0.002 \\
    Qwen2.5-1.5B     & Semantic-neighbour filler   & 0.012 & \textbf{0.154} \\
    Qwen2.5-1.5B     & Unrelated-repeat block      & 0.000 & 0.001 \\
    \midrule
    Qwen2.5-7B       & Full $N{=}30$ repeat        & 0.109 & 0.003 \\
    Qwen2.5-7B       & Truncated $N{=}3$ repeat    & 0.124 & 0.005 \\
    Qwen2.5-7B       & Deleted block               & 0.000 & 0.000 \\
    Qwen2.5-7B       & Random filler               & 0.001 & 0.003 \\
    Qwen2.5-7B       & Semantic-neighbour filler   & 0.013 & \textbf{0.164} \\
    Qwen2.5-7B       & Unrelated-repeat block      & 0.000 & 0.000 \\
    \midrule
    OLMo-2-7B        & Full $N{=}30$ repeat        & 0.042 & 0.006 \\
    OLMo-2-7B        & Truncated $N{=}3$ repeat    & 0.027 & 0.006 \\
    OLMo-2-7B        & Deleted block               & 0.000 & 0.000 \\
    OLMo-2-7B        & Random filler               & 0.001 & 0.002 \\
    OLMo-2-7B        & Semantic-neighbour filler   & 0.013 & \textbf{0.105} \\
    OLMo-2-7B        & Unrelated-repeat block      & 0.000 & 0.000 \\
    \bottomrule
  \end{tabular}
  \caption{Mean $P(\text{target})$ and mean $\Sigma P(\text{synonyms})$
    per (model, ablation condition) on the F0 displaced probe.
    Non-semantic conditions use the per-model single-token-valid set;
    the semantic-neighbour filler and the semantic-field mass are
    computed on the targets with curated semantic dictionaries. The
    semantic-neighbour filler is the only condition that raises the
    semantic-field mass substantially.}
  \label{tab:abl_full}
\end{table*}

\section{Frame-Pragmatics Ablation}
\label{app:framepragm}

\paragraph{Motivation and design.}
The displaced probe combines a repeated block with a topic-introducer
frame; \S\ref{sec:framepragm} asks whether the discourse pragmatics of
that frame, rather than repetition at a displaced slot, drives the
inverted-U. Holding the displaced readout position and the repeated
target block fixed, we vary only the frame's pragmatic class on the
four mechanism models, scoring $P(\text{target})$ exactly as in the
main displaced grid (per-model single-token targets,
$N \in \{0,1,3,5,10,20,30\}$, per-word drop with a $500$-resample
word-level bootstrap, seed $42$).

\paragraph{Frame classes.}
Four classes are compared. \emph{Topic-introducer} reuses the main
frames F0 (\codename{`The word I keep thinking about is'}) and F2
(\codename{`I want to mention'}). \emph{Neutral} uses existential
continuations with no novelty or anaphora pragmatics
(\codename{`I saw a'}, \codename{`There was a'}, \codename{`Look, a'}).
\emph{Anaphoric} frames demand the just-repeated word
(\codename{`That word again:'}, \codename{`The word repeated above
is'}, \codename{`To say it once more, the word is'}). \emph{Novelty}
frames demand a different word
(\codename{`On a completely different note, the word is'},
\codename{`Changing the subject, the new word is'},
\codename{`Now for something different, the word is'}). Per-class
drops average over the frames in the class.

\begin{figure*}[!htbp]
  \centering
  \includegraphics[width=\textwidth]{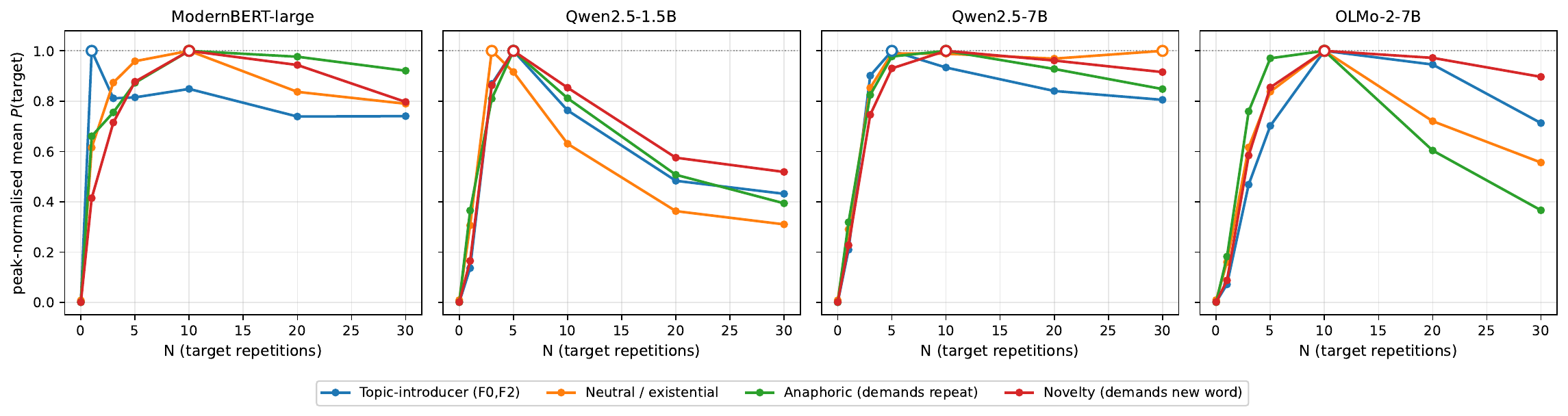}
  \caption{\textbf{Frame-pragmatics ablation, four models.}
    Peak-normalised mean $P(\text{target})$ on the displaced probe
    with readout position and repeated block fixed and only the
    frame's pragmatic class varied. Open dot marks each curve's peak
    over $N \ge 1$. The inverted-U persists across all four pragmatic
    classes, including anaphoric frames that demand the repeated word;
    per-(model, class) drops and CIs in Table~\ref{tab:framepragm}.}
  \label{fig:framepragm}
\end{figure*}

\paragraph{Result.}
All $16$ (model, class) cells show a positive per-word drop with
bootstrap CI excluding zero (Table~\ref{tab:framepragm}), and the
peak-normalised curves invert in $15$ of $16$
(Figure~\ref{fig:framepragm}). The anaphoric class keeps the
inverted-U in every model, and the novelty class (which a
discourse-novelty account predicts should produce the steepest
decline) is the largest-drop class in none of the four models. The
effect therefore does not require, and is not amplified by, a frame
that pragmatically favours a new word.

\begin{table*}[!htbp]
  \centering
  \footnotesize
  \setlength{\tabcolsep}{5pt}
  \begin{tabular}{l l r r r c c}
    \toprule
    \textbf{Model} & \textbf{Frame class} & $n_w$ & \textbf{drop} & \textbf{95\% CI} & \textbf{peak }$N$ & \textbf{inverts} \\
    \midrule
    ModernBERT-large & Topic-introducer (F0,F2) & 250 & $0.382$ & $[0.346,\,0.421]$ & 1 & yes \\
     & Neutral / existential & 250 & $0.371$ & $[0.332,\,0.412]$ & 10 & yes \\
     & Anaphoric (demands repeat) & 250 & $0.129$ & $[0.109,\,0.149]$ & 10 & yes \\
     & Novelty (demands new) & 250 & $0.312$ & $[0.281,\,0.341]$ & 10 & yes \\
    \midrule
    Qwen2.5-1.5B & Topic-introducer (F0,F2) & 256 & $0.713$ & $[0.681,\,0.744]$ & 5 & yes \\
     & Neutral / existential & 256 & $0.748$ & $[0.717,\,0.779]$ & 3 & yes \\
     & Anaphoric (demands repeat) & 256 & $0.719$ & $[0.673,\,0.760]$ & 5 & yes \\
     & Novelty (demands new) & 256 & $0.657$ & $[0.614,\,0.697]$ & 5 & yes \\
    \midrule
    Qwen2.5-7B & Topic-introducer (F0,F2) & 256 & $0.311$ & $[0.279,\,0.341]$ & 5 & yes \\
     & Neutral / existential & 256 & $0.194$ & $[0.166,\,0.221]$ & 30 & no \\
     & Anaphoric (demands repeat) & 256 & $0.213$ & $[0.190,\,0.237]$ & 10 & yes \\
     & Novelty (demands new) & 256 & $0.298$ & $[0.266,\,0.332]$ & 10 & yes \\
    \midrule
    OLMo-2-7B & Topic-introducer (F0,F2) & 256 & $0.426$ & $[0.386,\,0.465]$ & 10 & yes \\
     & Neutral / existential & 256 & $0.483$ & $[0.443,\,0.518]$ & 10 & yes \\
     & Anaphoric (demands repeat) & 256 & $0.661$ & $[0.638,\,0.686]$ & 10 & yes \\
     & Novelty (demands new) & 256 & $0.284$ & $[0.245,\,0.323]$ & 10 & yes \\
    \bottomrule
  \end{tabular}
  \caption{\textbf{Frame-pragmatics ablation.} Per-word drop $(\text{peak}_w - P(N{=}30)_w)/\text{peak}_w$ (peak over $N\ge1$) on the displaced probe with the readout position and repeated block held fixed and only the frame's pragmatic class varied, four-model subset. $n_w$ is the per-model single-token target count; $95\%$ bootstrap CI ($500$ resamples, seed $42$, word-level). \textbf{inverts} marks whether the population-mean curve peaks before $N{=}30$. All $16$ cells show a positive per-word drop with CI excluding zero; the anaphoric class, whose frames demand the repeated word, keeps the inverted-U in every model, and the novelty class is the largest-drop class in none of the four models. Frame strings are listed in the appendix text above.}
  \label{tab:framepragm}
\end{table*}

\section{Cross-Lingual Panel}
\label{app:multiling}

\paragraph{Checkpoints and setup details.}
The four multilingual checkpoints used in \S\ref{sec:lang} are
\codename{FacebookAI/xlm-roberta-base} (MLM, $279$M) and
\codename{FacebookAI/xlm-roberta-large} (MLM,
$561$M)~\citep{conneau2020xlmr}; \codename{Qwen/Qwen2.5-1.5B} and
\codename{Qwen/Qwen2.5-7B}~\citep{qwen2025qwen25} (CLMs). Word lists
are hand-curated single-token nouns of length $\geq 3$, balanced for
concreteness/frequency where possible; each candidate is filtered
against the model's own tokeniser using the same leading-space-first
rule as the English panel. Cross-lingual frames F0--F2 are
hand-translated to idiomatic equivalents of the English F0--F2
templates; F3 and F4 are not translated, so the cross-lingual panel
is evaluated on F0--F2 only.

\paragraph{Per-(model, language, frame) results.}
Table~\ref{tab:multilingual} reports peak $N$, peak $P$, drop, and
$95\%$ bootstrap CI for every cell. Mean-curve peaks land at
$N{=}1$ for XLM-R-base, $N{=}3$--$10$ for both Qwen2.5 sizes, and
split between $N{=}1$ and $N{=}30$ for XLM-R-large. The twelve
(model, language, frame) cells whose mean curve peaks at $N{=}30$ are
five Qwen2.5 cells (1.5B on German F2, French F1, French F2; 7B on
German F2 and French F2) and seven XLM-R-large cells (all six of its
non-German cells plus German F0); all keep positive per-word drops.
By language, F0 drops span $0.308$ (XLM-R-large) to $0.853$
(Qwen2.5-1.5B) in Spanish, $0.354$--$0.814$ in German,
$0.443$--$0.823$ in French, and $0.435$ (Qwen2.5-7B) to $0.481$
(Qwen2.5-1.5B) in Chinese.

\begin{table*}[!htbp]
  \centering
  \footnotesize
  \setlength{\tabcolsep}{5pt}
  \begin{tabular}{l c c c c c c c}
    \toprule
    \textbf{Model} & \textbf{Lang.} & \textbf{Frame} & $n_w$ & \textbf{Peak $N$} & \textbf{Peak $P$} & \textbf{Drop} & \textbf{95\,\% CI of drop} \\
    \midrule
    xlm-roberta-base & German & F0 & 46 & 1 & 0.784 & 0.515 & $[0.412,\,0.618]$ \\
    xlm-roberta-base & German & F1 & 46 & 1 & 0.302 & 0.561 & $[0.465,\,0.662]$ \\
    xlm-roberta-base & German & F2 & 46 & 20 & 0.006 & 0.502 & $[0.410,\,0.583]$ \\
    xlm-roberta-base & Spanish & F0 & 49 & 1 & 0.492 & 0.541 & $[0.431,\,0.636]$ \\
    xlm-roberta-base & Spanish & F1 & 49 & 1 & 0.210 & 0.737 & $[0.665,\,0.803]$ \\
    xlm-roberta-base & Spanish & F2 & 49 & 20 & 0.062 & 0.539 & $[0.442,\,0.630]$ \\
    xlm-roberta-base & French & F0 & 40 & 1 & 0.442 & 0.823 & $[0.731,\,0.903]$ \\
    xlm-roberta-base & French & F1 & 40 & 10 & 0.063 & 0.573 & $[0.447,\,0.686]$ \\
    xlm-roberta-base & French & F2 & 40 & 1 & 0.088 & 0.543 & $[0.432,\,0.651]$ \\
    \midrule
    xlm-roberta-large & German & F0 & 46 & 30 & 0.265 & 0.354 & $[0.257,\,0.455]$ \\
    xlm-roberta-large & German & F1 & 46 & 1 & 0.037 & 0.561 & $[0.455,\,0.671]$ \\
    xlm-roberta-large & German & F2 & 46 & 1 & 0.009 & 0.869 & $[0.793,\,0.929]$ \\
    xlm-roberta-large & Spanish & F0 & 49 & 30 & 0.207 & 0.308 & $[0.219,\,0.389]$ \\
    xlm-roberta-large & Spanish & F1 & 49 & 30 & 0.045 & 0.372 & $[0.273,\,0.479]$ \\
    xlm-roberta-large & Spanish & F2 & 49 & 30 & 0.002 & 0.749 & $[0.656,\,0.830]$ \\
    xlm-roberta-large & French & F0 & 40 & 30 & 0.208 & 0.443 & $[0.335,\,0.555]$ \\
    xlm-roberta-large & French & F1 & 40 & 30 & 0.012 & 0.115 & $[0.062,\,0.181]$ \\
    xlm-roberta-large & French & F2 & 40 & 30 & 0.004 & 0.639 & $[0.508,\,0.756]$ \\
    \midrule
    Qwen2.5-1.5B & German & F0 & 28 & 5 & 0.192 & 0.814 & $[0.718,\,0.893]$ \\
    Qwen2.5-1.5B & German & F1 & 28 & 5 & 0.036 & 0.624 & $[0.535,\,0.714]$ \\
    Qwen2.5-1.5B & German & F2 & 28 & 30 & 0.142 & 0.112 & $[0.034,\,0.214]$ \\
    Qwen2.5-1.5B & Spanish & F0 & 34 & 5 & 0.199 & 0.853 & $[0.785,\,0.909]$ \\
    Qwen2.5-1.5B & Spanish & F1 & 34 & 5 & 0.003 & 0.579 & $[0.479,\,0.669]$ \\
    Qwen2.5-1.5B & Spanish & F2 & 34 & 5 & 0.001 & 0.684 & $[0.600,\,0.761]$ \\
    Qwen2.5-1.5B & French & F0 & 29 & 3 & 0.105 & 0.777 & $[0.686,\,0.861]$ \\
    Qwen2.5-1.5B & French & F1 & 29 & 30 & 0.002 & 0.481 & $[0.361,\,0.608]$ \\
    Qwen2.5-1.5B & French & F2 & 29 & 30 & 0.001 & 0.568 & $[0.464,\,0.680]$ \\
    Qwen2.5-1.5B & Chinese & F0 & 49 & 3 & 0.007 & 0.481 & $[0.411,\,0.549]$ \\
    Qwen2.5-1.5B & Chinese & F1 & 49 & 3 & 0.007 & 0.367 & $[0.287,\,0.451]$ \\
    Qwen2.5-1.5B & Chinese & F2 & 49 & 10 & 0.007 & 0.384 & $[0.298,\,0.472]$ \\
    \midrule
    Qwen2.5-7B & German & F0 & 28 & 10 & 0.229 & 0.479 & $[0.366,\,0.583]$ \\
    Qwen2.5-7B & German & F1 & 28 & 5 & 0.040 & 0.610 & $[0.527,\,0.706]$ \\
    Qwen2.5-7B & German & F2 & 28 & 30 & 0.087 & 0.117 & $[0.047,\,0.203]$ \\
    Qwen2.5-7B & Spanish & F0 & 34 & 5 & 0.192 & 0.732 & $[0.646,\,0.808]$ \\
    Qwen2.5-7B & Spanish & F1 & 34 & 5 & 0.010 & 0.448 & $[0.374,\,0.527]$ \\
    Qwen2.5-7B & Spanish & F2 & 34 & 20 & 0.001 & 0.305 & $[0.222,\,0.400]$ \\
    Qwen2.5-7B & French & F0 & 29 & 5 & 0.152 & 0.486 & $[0.392,\,0.587]$ \\
    Qwen2.5-7B & French & F1 & 29 & 5 & 0.007 & 0.378 & $[0.291,\,0.452]$ \\
    Qwen2.5-7B & French & F2 & 29 & 30 & 0.002 & 0.202 & $[0.112,\,0.306]$ \\
    Qwen2.5-7B & Chinese & F0 & 49 & 5 & 0.027 & 0.435 & $[0.386,\,0.483]$ \\
    Qwen2.5-7B & Chinese & F1 & 49 & 3 & 0.012 & 0.575 & $[0.504,\,0.639]$ \\
    Qwen2.5-7B & Chinese & F2 & 49 & 5 & 0.027 & 0.387 & $[0.321,\,0.455]$ \\
    \bottomrule
  \end{tabular}
  \caption{Cross-lingual results. Four open-access multilingual
    models (XLM-R-base, XLM-R-large, Qwen2.5-1.5B, Qwen2.5-7B) on
    Spanish, Chinese, German, and French single-token noun lists, with
    three hand-translated frames per language. \textit{Peak $N$} and
    \textit{Peak $P$} are taken from the word-averaged mean curve;
    \textit{Drop} is the per-word mean of $(\text{peak}_w -
    P(N{=}30)_w)/\text{peak}_w$, with bootstrap $95\,\%$ CI over the
    per-(model, language) word population. XLM-R Chinese is excluded
    for both base and large because XLM-R's BPE leaves no Chinese
    single-token candidates. All $42$ evaluated cells show a positive
    drop with CI excluding zero; $30$ of $42$ also have mean-curve
    peak before $N{=}30$.}
  \label{tab:multilingual}
\end{table*}

\paragraph{Panel scope.}
The four-model panel spans both training objectives, two tokeniser
families (XLM-R SentencePiece and Qwen BPE), and four target
languages.

\section{Mechanism Probes: Per-$N$ and Layer-wise}
\label{app:layers}

\paragraph{Per-$N$ attention table.}
Table~\ref{tab:mech_attn} reports per-target-token attention at the
displaced slot in the last layer (averaged across heads and target
words), the $1/N$ reference anchored at the smallest evaluated $N$,
and the total attention to the repetition block. In three of four
mechanism models the total block budget grows monotonically with
$N$; ModernBERT-large is the exception, with a roughly flat block
budget.

\begin{table*}[!htbp]
  \centering
  \footnotesize
  \setlength{\tabcolsep}{5pt}
  \begin{tabular}{l c c c c}
    \toprule
    \textbf{Model} & $N$ & \textbf{Per-token} & \textbf{$1/N$ ref.} & \textbf{Total to block} \\
    \midrule
    Qwen2.5-1.5B  & 3  & 0.054 & 0.054 & 0.109 \\
    Qwen2.5-1.5B  & 5  & 0.035 & 0.033 & 0.141 \\
    Qwen2.5-1.5B  & 10 & 0.026 & 0.016 & 0.235 \\
    Qwen2.5-1.5B  & 20 & 0.020 & 0.008 & 0.383 \\
    Qwen2.5-1.5B  & 30 & \textbf{0.016} & \textbf{0.005} & 0.471 \\
    \midrule
    Qwen2.5-7B    & 3  & 0.031 & 0.031 & 0.061 \\
    Qwen2.5-7B    & 5  & 0.040 & 0.018 & 0.159 \\
    Qwen2.5-7B    & 10 & 0.024 & 0.009 & 0.214 \\
    Qwen2.5-7B    & 20 & 0.013 & 0.005 & 0.246 \\
    Qwen2.5-7B    & 30 & \textbf{0.010} & \textbf{0.003} & 0.292 \\
    \midrule
    OLMo-2-7B     & 3  & 0.015 & 0.015 & 0.030 \\
    OLMo-2-7B     & 5  & 0.011 & 0.009 & 0.045 \\
    OLMo-2-7B     & 10 & 0.007 & 0.004 & 0.067 \\
    OLMo-2-7B     & 20 & 0.006 & 0.002 & 0.122 \\
    OLMo-2-7B     & 30 & \textbf{0.008} & \textbf{0.002} & 0.242 \\
    \midrule
    ModernBERT-large & 3  & 0.044 & 0.044 & 0.089 \\
    ModernBERT-large & 5  & 0.026 & 0.027 & 0.105 \\
    ModernBERT-large & 10 & 0.017 & 0.013 & 0.150 \\
    ModernBERT-large & 20 & 0.008 & 0.007 & 0.157 \\
    ModernBERT-large & 30 & \textbf{0.005} & \textbf{0.004} & 0.143 \\
    \bottomrule
  \end{tabular}
  \caption{Per-target-token attention at the displaced slot in the
    last layer. ``$1/N$ ref.'' is the budget-dilution baseline
    anchored at the smallest evaluated $N{=}3$; bold entries
    highlight the $N{=}30$ comparison. In three of four models the
    total block budget grows with $N$; ModernBERT-large's stays
    roughly flat.}
  \label{tab:mech_attn}
\end{table*}

\begin{figure*}[!htbp]
  \centering
  \includegraphics[width=\textwidth]{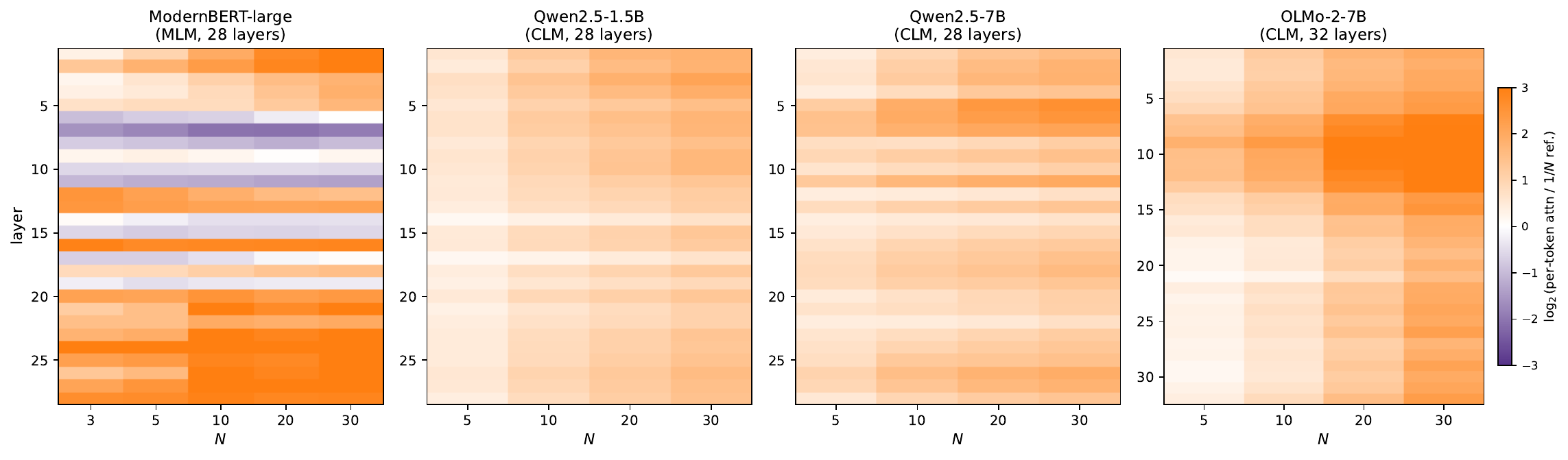}
  \caption{\textbf{Layer-wise per-target-token attention vs the
    per-layer $1/N$ reference}, displaced slot, four mechanism
    models on the $50$-word stratified subset. Cells show
    $\log_{2}$(per-token attn / $1/N$ ref.\ at smallest $N$). Purple
    = super-dilution (per-token attention falls below the
    budget-dilution baseline); orange = sub-dilution (the two
    endpoints differ in luminance for grayscale readers).
    ModernBERT-large super-dilutes in its upper layers; Qwen2.5-1.5B,
    Qwen2.5-7B, and OLMo-2-7B sit predominantly above their $1/N$
    references across the stack.}
  \label{fig:layerattn}
\end{figure*}

\paragraph{Layer-wise per-token attention.}
Figure~\ref{fig:layerattn} extends the last-layer analysis above to
every layer of the four mechanism models on the $50$-word stratified
subset.

\section{LAMA-Style Pilot}
\label{app:lama}

\paragraph{Setup.}
Three answer-prompt templates of the form
\codename{<prefix> the answer is <target> . <block> . <suffix> the
answer is} (A0--A2 paraphrase the prefix and suffix) are evaluated
on the four mechanism models. The middle \codename{<block>} is
$N \in \{1,3,5,10,20,30\}$ tokens long, drawn from one of four
conditions: \codename{target\_repeat} (the target itself),
\codename{random\_repeat} (random neutral nouns),
\codename{unrelated\_repeat} (\codename{`table'}), or
\codename{semantic\_repeat} (semantic neighbours). Each (model,
template, condition, $N$) cell is run on a $50$-word stratified
subset of that model's single-token-valid targets; per-word
probabilities are averaged across the three templates.

\paragraph{Target-repeat curves.}
Table~\ref{tab:lama_curves} reports the mean $P(\text{target})$ at
each $N$ for the \codename{target\_repeat} block (the directly
comparable condition), and the per-word drop
$(\text{peak}_w - P(N{=}30)_w)/\text{peak}_w$ with $95\%$ bootstrap
CI ($500$ resamples, seed $42$). Three of four models replicate the
displaced attenuation with CI excluding zero; the magnitude is
smaller than the headline displaced effect because the pre-stated
answer provides a second retrieval route.

\begin{table*}[!htbp]
  \centering
  \footnotesize
  \setlength{\tabcolsep}{4pt}
  \begin{tabular}{l c c c c c c c c}
    \toprule
    \textbf{Model} & $N{=}1$ & $N{=}3$ & $N{=}5$ & $N{=}10$ & $N{=}20$ & $N{=}30$ & \textbf{drop} & \textbf{95\% CI} \\
    \midrule
    ModernBERT-large & $0.622$ & $0.585$ & $0.636$ & $0.727$ & $\mathbf{0.744}$ & $0.740$ & $0.058$ & $[0.026,\,0.097]$ \\
    Qwen2.5-1.5B     & $0.356$ & $0.432$ & $\mathbf{0.450}$ & $0.405$ & $0.406$ & $0.414$ & $0.136$ & $[0.099,\,0.171]$ \\
    Qwen2.5-7B       & $0.446$ & $0.518$ & $0.589$ & $0.621$ & $\mathbf{0.621}$ & $0.596$ & $0.116$ & $[0.078,\,0.165]$ \\
    OLMo-2-7B        & $0.270$ & $0.347$ & $0.417$ & $0.496$ & $0.565$ & $\mathbf{0.622}$ & $0.002$ & $[0.000,\,0.005]$ \\
    \bottomrule
  \end{tabular}
  \caption{LAMA-style pilot: mean $P(\text{target})$ across $50$
    stratified words and three answer-prompt templates, for the
    \codename{target\_repeat} block at each $N$. Bold entries mark
    each model's peak. Per-word drop is
    $(\text{peak}_w - P(N{=}30)_w)/\text{peak}_w$ with $95\%$
    bootstrap CI. Three of four models show a positive drop with CI
    excluding zero; OLMo-2-7B's mean curve does not invert through
    $N{=}30$ and its CI lower bound rounds to zero at the reported
    precision (the unrounded $2.5\%$ quantile is a small positive
    number, not strictly $0$).}
  \label{tab:lama_curves}
\end{table*}

\paragraph{Control conditions.}
The \codename{unrelated\_repeat} and \codename{random\_repeat}
blocks (which break the target match) produce baseline
$P(\text{target})$ in the $0.13$--$0.45$ range across models without
the inverted-U signature, confirming that the
\codename{target\_repeat} effect is target-specific rather than a
generic position artefact. The \codename{semantic\_repeat} block
falls between these and the target-repeat block. Full per-condition
curves are in the released data.

\section{Reproducibility}
\label{app:repro}

\paragraph{Code, data, and artefact documentation.}
The complete implementation, reproducibility scripts, released
result files, and artefact documentation are available at
\url{https://github.com/henryhyw/when-more-becomes-less}.

\paragraph{Artefacts, licenses, and intended use.}
The released package contains the analysis code, reproducibility
scripts, target lists, per-model output tables, figure-generation
scripts, and result files needed to audit the claims in the paper.
The released code is MIT-licensed; pretrained model checkpoints and
other third-party artefacts retain their original licenses and terms.
The intended use of the released artefacts is research reproduction,
inspection, and extension of the two-probe diagnostic, not deployment
as a model-control or safety system.

\paragraph{Data, human subjects, and risk profile.}
The study uses no human subjects, annotators, private user data, or
newly collected natural-language documents. The English panel is a
hand-curated list of common single-token nouns filtered per
tokeniser; multilingual and mechanism probes use fixed prompt
templates, translated frames, and model-generated probability
outputs. We did not intentionally include personally identifying
information or offensive content. The main risk is methodological
over-interpretation: the results show that repetition count and
readout position interact in controlled cloze probes, not that
truncation or repetition shaping is a general-purpose intervention
for open-ended generation.

\paragraph{Pipeline.}
The full analysis pipeline is invoked through a single notebook
that runs target-set construction, displaced evaluations, ablations,
mechanism probes, figure rendering, and verification. The per-stage
script-to-output mapping is documented in the reproducibility
README. The mechanism and ablation probes for the four high-cost
models are scoped to finish within $\approx 12$ hours on a single
A100 80GB.

\paragraph{Hardware.}
The full panel runs end-to-end on a single A100 80GB GPU in
\texttt{bfloat16}. Smaller models (under $4$B) finish in under ten
minutes each; Qwen2.5-7B and the OLMo-2-7B mechanism battery take
roughly $30$--$60$ minutes; Qwen2.5-14B (displaced grid only) takes
roughly ninety minutes.

\paragraph{Model checkpoints.}
All models are loaded from public HuggingFace checkpoints with no
access tokens required. The exact checkpoint identifiers are
\codename{answerdotai/ModernBERT-base},
\codename{answerdotai/ModernBERT-large},
\codename{Qwen/Qwen2.5-0.5B},
\codename{Qwen/Qwen2.5-1.5B},
\codename{Qwen/Qwen2.5-3B},
\codename{Qwen/Qwen2.5-7B},
\codename{Qwen/Qwen2.5-7B-Instruct},
\codename{Qwen/Qwen2.5-14B},
\codename{HuggingFaceTB/SmolLM2-360M},
\codename{HuggingFaceTB/SmolLM2-1.7B},
\codename{microsoft/Phi-3.5-mini-instruct},
\codename{allenai/OLMo-2-0425-1B},
and \codename{allenai/OLMo-2-1124-7B}.

\paragraph{Random seeds.}
All bootstrap resampling uses seed $42$ ($500$ resamples). All
word-list stratifications and ablation samplings use seed $0$. Model
forward passes are deterministic.

\paragraph{Pinned dependencies.}
Exact versions of PyTorch, HuggingFace Transformers (>= $4.48$),
pandas, numpy, scipy, statsmodels, matplotlib, and \codename{wordfreq}
are pinned at the snapshot used to produce the reference outputs.

\paragraph{Verification.}
A verification script enumerates the expected output manifest
(per-model target-word CSVs, displaced/ablation parquet tables,
mechanism CSVs, figure files, and the compiled PDF) and reports
per-file presence and size. The reference
reproduction run passes the manifest in full.

\end{document}